\documentclass{article}
\usepackage{microtype}
\usepackage{graphicx}
\usepackage{subfigure}
\usepackage{booktabs} 
\usepackage{caption}  
\usepackage[dvipsnames,table]{xcolor}

\usepackage{hyperref}

\usepackage[accepted]{icml2025} 

\usepackage{amsmath}
\usepackage{amssymb}
\usepackage{mathtools}
\usepackage{amsthm}
\usepackage{wrapfig} 
\usepackage{amsmath}
\usepackage{amsfonts}
\usepackage{amssymb}
\usepackage{bm}
\usepackage{multirow}
\usepackage{xspace}  
\usepackage[normalem]{ulem}

\newcommand{\up}[1]{\textcolor{OliveGreen}{\small \ $\uparrow${#1}}}
\newcommand{\downbad}[1]{\textcolor{Maroon}{\small \ $\downarrow${#1}}}

\newcommand{\normup}[1]{\textcolor{OliveGreen}{\normalsize \ $\uparrow${#1}}}

\newcommand{\model}{\textsc{AdaptVis}\xspace}
\newcommand{\vlm}{VLMs\xspace}
\newcommand{\scal}{\textsc{ScalingVis}\xspace}

\definecolor{adaptive}{rgb}{0.9, 0.92, 1.0}
\definecolor{scal}{rgb}{1.0, 0.9, 0.92}
\definecolor{ood}{rgb}{1.0, 1.0, 0.92}
\definecolor{closed-loop}{RGB}{233, 196, 107}
\definecolor{scaling}{RGB}{130, 178,154}
\definecolor{global-optimal}{RGB}{246, 111, 105}
\definecolor{efficirency}{RGB}{033, 158,188}
\definecolor{strings}{RGB}{220, 20, 60}

\def\mA{{\bm{A}}}

\def\mK{{\bm{K}}}

\def\mM{{\bm{M}}}
\def\mN{{\bm{N}}}

\def\mQ{{\bm{Q}}}

\def\mV{{\bm{V}}}
\def\mW{{\bm{W}}}
\def\mX{{\bm{X}}}

\usepackage[toc,page,header]{appendix}
\usepackage{minitoc}

\usepackage[capitalize,noabbrev]{cleveref}

\theoremstyle{plain}

\theoremstyle{definition}

\theoremstyle{remark}

\usepackage[textsize=tiny]{todonotes}
\DeclareUnicodeCharacter{FF1F}{?}

\icmltitlerunning{Why Is Spatial Reasoning Hard for VLMs? An Attention Mechanism Perspective on Focus Areas}

\begin{document}

\twocolumn[
\icmltitle{Why Is Spatial Reasoning Hard for VLMs? \\An Attention Mechanism Perspective on Focus Areas}

\begin{icmlauthorlist}
  \icmlauthor{Shiqi Chen}{cityu}%
  \icmlauthor{Tongyao Zhu}{nus}%
  \icmlauthor{Ruochen Zhou}{cityu}%
  \icmlauthor{Jinghan Zhang}{hkust}%
  \icmlauthor{Siyang Gao}{cityu}%
  \icmlauthor{Juan Carlos Niebles}{stanford,salesforce}%
  \icmlauthor{Mor Geva}{telaviv}%
  \icmlauthor{Junxian He}{hkust}%
  \icmlauthor{Jiajun Wu}{stanford}%
  \icmlauthor{Manling Li}{stanford,northwestern}%
\end{icmlauthorlist}

\icmlaffiliation{cityu}{City University of Hong Kong}
\icmlaffiliation{nus}{National University of Singapore}
\icmlaffiliation{hkust}{Hong Kong University of Science and Technology}
\icmlaffiliation{stanford}{Stanford University}
\icmlaffiliation{salesforce}{Salesforce Research}
\icmlaffiliation{telaviv}{Tel Aviv University}
\icmlaffiliation{northwestern}{Northwestern University}

\icmlcorrespondingauthor{Shiqi Chen}{schen438-c@my.cityu.edu.hk}
\icmlcorrespondingauthor{Manling Li}{manling.li@northwestern.edu}

\icmlkeywords{Machine Learning, ICML}
\vskip 0.1in
\vskip 0.2in
]



\printAffiliationsAndNotice{}

\newcommand\encircle[2][]{\tikz[overlay]\node[fill=blue!20,inner sep=2pt, anchor=text, rectangle, rounded corners=1.5mm,#1] {#2};\phantom{#2}}

\begin{figure*}[t]
    \centering
    \includegraphics[width=\textwidth]{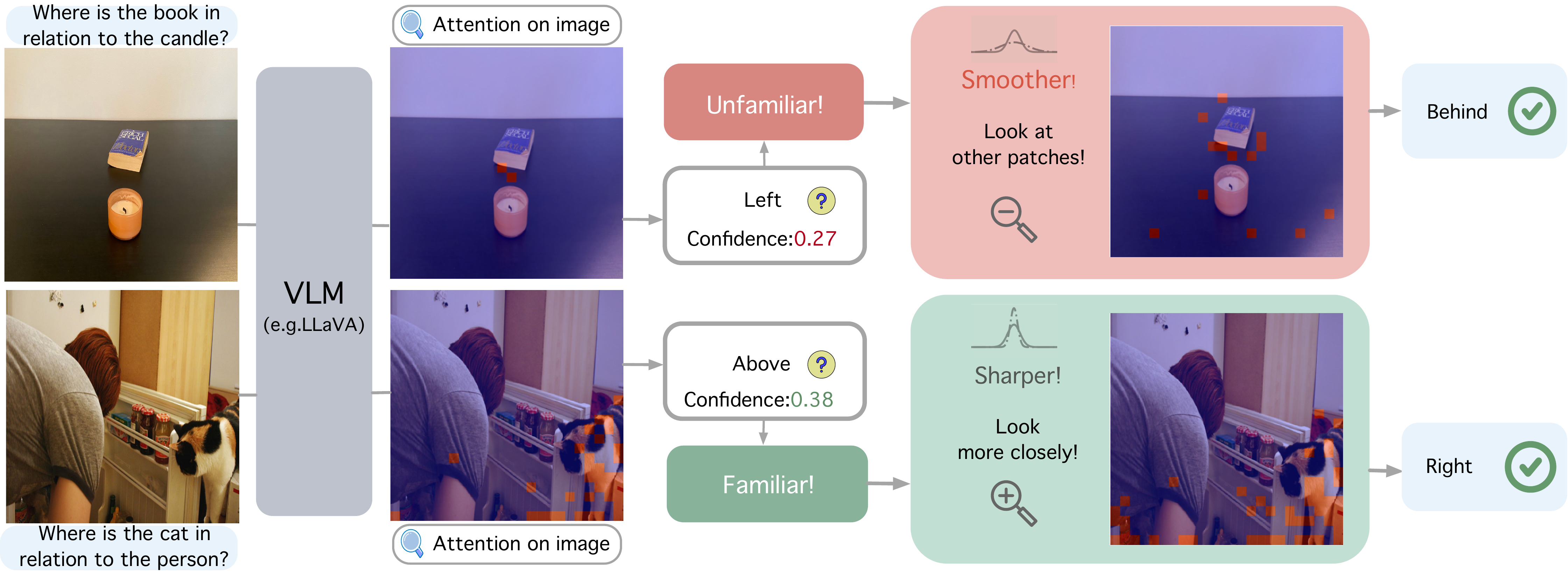}
    \caption{The framework of \model. We adaptively intervene in the temperature of the attention logits of the image tokens. Top: For generations with \textbf{low} confidence, we \textbf{smoothen} the attention distribution to broaden the context window for better concentration on the correct objects. Bottom: For generations with \textbf{high} confidence, we trust the attention pattern and \textbf{sharpen} the attention distribution.}
    \label{fig:main-figure}
\end{figure*}

\begin{abstract} 
Large Vision Language Models (\vlm) have long struggled with spatial reasoning tasks. Surprisingly, even simple spatial reasoning tasks, such as recognizing ``under'' or ``behind'' relationships between only two objects, pose significant challenges for current \vlm. 
In this work, we study the spatial reasoning challenge from the lens of~\textbf{mechanistic interpretability}, diving into the model's internal states to examine the interactions between image and text tokens. 
By tracing attention distribution over the image throughout intermediate layers, we observe that successful spatial reasoning correlates strongly with the model's ability to align its attention distribution with actual object locations, particularly differing between familiar and unfamiliar spatial relationships.
Motivated by these findings, we propose \model based on inference-time confidence scores to sharpen the attention on highly relevant regions when confident, while smoothing and broadening the attention window to consider a wider context when confidence is lower.
This training-free decoding method shows significant improvement (e.g., up to a 50 absolute point improvement) on spatial reasoning benchmarks such as WhatsUp and VSR with negligible cost.  We make code and data publicly available for research purposes at {\small \url{https://github.com/shiqichen17/AdaptVis}}. 
\end{abstract}
\section{Introduction}
Despite rapid advancements in Large Vision-Language Models (VLMs), a significant deficiency persists, i.e., their struggle with vision-centric abilities \citep{gao2023g,kamath-etal-2023-whats,tong2024cambrian,chen2024spatialvlm}. 
This limitation is particularly notable in spatial reasoning given the simplicity of the task. Spatial reasoning involves inferring basic relationships between just two objects, such as ``\textit{left}'', ``\textit{right}'', ``\textit{above}'', ``\textit{below}'', ``\textit{behind}'', or ``\textit{front}'', as shown in Figure~\ref{fig:main-figure}.
For example, given the image with a book ``\textit{behind}'' the candle in Figure~\ref{fig:main-figure}, \vlm describe the book as being ``\textit{left}'' of the candle.  
This error is not an isolated incident but a frequent, recurring pattern, highlighting a fundamental bottleneck in how VLMs process visual-centric information.

Recent studies have probed the limitations of vision encoders like CLIP \citep{radford2021learning} in VLMs' vision processing 
\citep{tong2024eyes,tong2024cambrian}, yet a critical aspect remains underexplored: how vision and text tokens interact within the model's internal states to construct geometric understanding. 
Specifically, the model must simultaneously identify objects while maintaining awareness of the broader geometric context, grasping the complex geometric relationships between them. This geometric understanding fundamentally manifests in how models distribute their attention across the visual tokens. This unique challenge makes spatial reasoning an ideal lens for studying how VLMs internally process vision-centric information.

Therefore, we open up the black box of VLMs and examine their internal mechanisms through a suite of carefully designed spatial reasoning tasks. By looking into the attention distributions, we can systematically investigate how vision and text tokens interact to construct, or fail to construct, accurate spatial understanding. 
Our investigation begins with a key observation: despite image tokens comprising around 90\% of the input sequence, they receive only about 10\% of the model's attention. This significant imbalance suggests that textual priors often overshadow visual evidence, explaining VLMs' struggles with vision-centric tasks.

While a straightforward solution might seem to be simply increasing attention to visual tokens, our deeper analysis reveals that the challenge lies not just in the quantity but in the geometric distribution of visual attention. By examining attention patterns across model layers, we observe a pattern in VLMs' attention behavior: their initial attention distribution often reflects learned priors that may or may not align with the actual geometric distribution over the image. This led us to a key insight: rather than accepting this initial distribution, we can dynamically intervene based on the model's self-belief, which is measured using its confidence score.
As shown in Figure~\ref{fig:main-figure}, when the model exhibits high confidence in its spatial reasoning (as measured by generation probability), we sharpen its attention distribution to strengthen the focus on its current beliefs. Conversely, when confidence is low, we smooth the attention distribution to encourage the exploration of alternative spatial relationships. 

We call this confidence-guided attention intervention \textsc{AdpatVis}, which proves remarkably effective while remaining computationally efficient. In experiments across diverse spatial reasoning benchmarks, including WhatsUp \citep{kamath-etal-2023-whats} and VSR \citep{Liu2022VisualSR}, our approach achieves substantial improvements of up to 50\% points. These gains are observed across both synthetic datasets with clean backgrounds and real-world images with complex scenes, demonstrating the robustness of our method.

Our visualization of this intervention strategy reveals its underlying mechanism: by dynamically adjusting attention patterns, we effectively guide the model's focus to better align with actual object locations and their spatial relationships. This suggests that successful spatial reasoning isn't just about having the right attention mechanism, but about having the right confidence to know when to trust or question one's initial spatial understanding.

\begin{figure*}[t!]
\centering

\begin{minipage}{0.55\textwidth}
    \centering
    \small
    \resizebox{1.0\linewidth}{!}{
    \begin{tabular}{l|cc|cc|cc}
    \toprule
    \textbf{Dataset} & \multicolumn{2}{c|}{\textbf{Num of Options}} & \multicolumn{2}{c}{\textbf{Num of Objects}} & \multicolumn{2}{c}{\textbf{Source}} \\
    \cmidrule(lr){2-3} \cmidrule(lr){4-5} \cmidrule(lr){6-7}
     & \textbf{4 options} & \textbf{6 options} & \textbf{1 object} & \textbf{2 objects}& \textbf{Syn} & \textbf{Real} \\
    \midrule
    Cont\_A  & \checkmark &   &  & \checkmark  &\checkmark&\\
    Cont\_B  & \checkmark &   &  & \checkmark  &\checkmark&  \\
    VG\_one        &   & \checkmark &   \checkmark&  && \checkmark\\
    VG\_two        &   & \checkmark &   & \checkmark&& \checkmark \\
    COCO\_one      & \checkmark  &  &  \checkmark & && \checkmark \\
    COCO\_two      & \checkmark &   &  & \checkmark && \checkmark \\
    \bottomrule
    \end{tabular}}
\end{minipage}
\hfill
\begin{minipage}{0.4\textwidth}
    \centering
    \includegraphics[width=\linewidth]{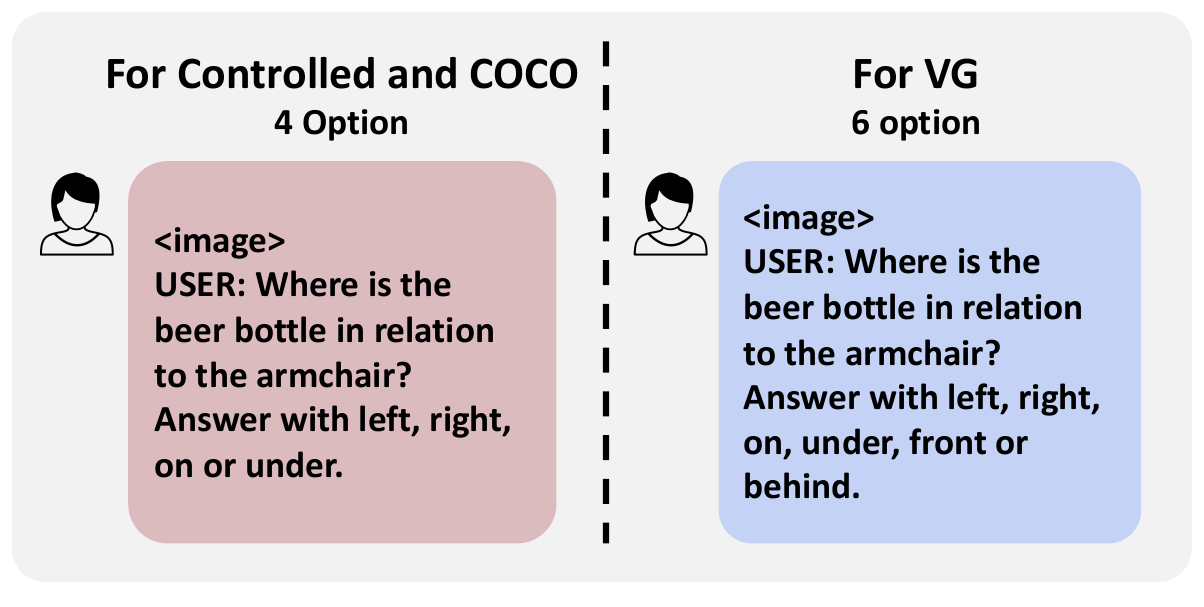}
    \label{fig:prompts}
\end{minipage}
\vspace{-4mm}
\caption{Left: Choice counts, object counts and data source by subset in WhatsUp (“Syn” = Synthetic, “Real” = Real). Right: Evaluation prompts we use in evaluation.}
\label{tab:dataset}
\end{figure*}
\section{Preliminary on VLMs}
\label{sec:background}
We center our analysis on investigating how \vlm distribute their attention over image tokens, aiming to gain deeper insights into spatial reasoning errors.

\paragraph{Notation} 
Large \vlm like LLaVA~\citep{liu2024visual} consist of three components: a visual encoder like CLIP~\citep{radford2021learning}, a pretrained language model, and a projector to connect them. The visual encoder functions as a perception tool to ``see" the image, while the image information is processed through a projector to be mapped into the token space. The LLM is often based on the transformer architecture, consisting of $L$ layers stacked together. Each layer consists of two major components: a Multi-Head Attention (MHA) module, followed by a feed-forward network. For each layer $l$, given the input $\mX \in \mathbb{R}^{n \times d}$ (where $n$ is the number of tokens and $d$ is the embedding dimension), MHA performs self-attention in each head $N_h$. The output is a concatenation of all heads' outputs: 
$
{\text{MHA}}^{(l)}(\mX) = \text{Concat}\left(\mN_h^{(l,1)}, \dots, \mN_h^{(l,H)}\right)\mW_o.
$
Here $H$ is the number of heads; $\mN_h$ is the output of head $N_h$: 

{\small
\begin{equation}
\mN_h^{(l,h)}  =  \text{Softmax}(\mA^{(l,h)})\mV =  \text{Softmax}\left(\frac{\mQ\mK^\top}{\sqrt{d_h}} + \mM\right)\mV,
\end{equation}
}\\
where attention logits $\mA^{(l,h)}$ are computed via 
$\mQ = \mX\mW_{q_h},\mK = \mX\mW_{k_h}, \mV = \mX\mW_{v_h}$, and $\mW_{q_h}, \mW_{k_h}, \mW_{v_h} \in \mathbb{R}^{d \times d_h}$ are learnable projection matrix of the head $N_h$.  
A causal mask $\mM_{ij} = 0$ if $i \geq j$, and $-\infty$ otherwise, prevents tokens from attending to future tokens.

\section{Text-Vision Attention Interactions}
\begin{figure}[t!]
    \centering
    \includegraphics[width=0.44\textwidth]
    {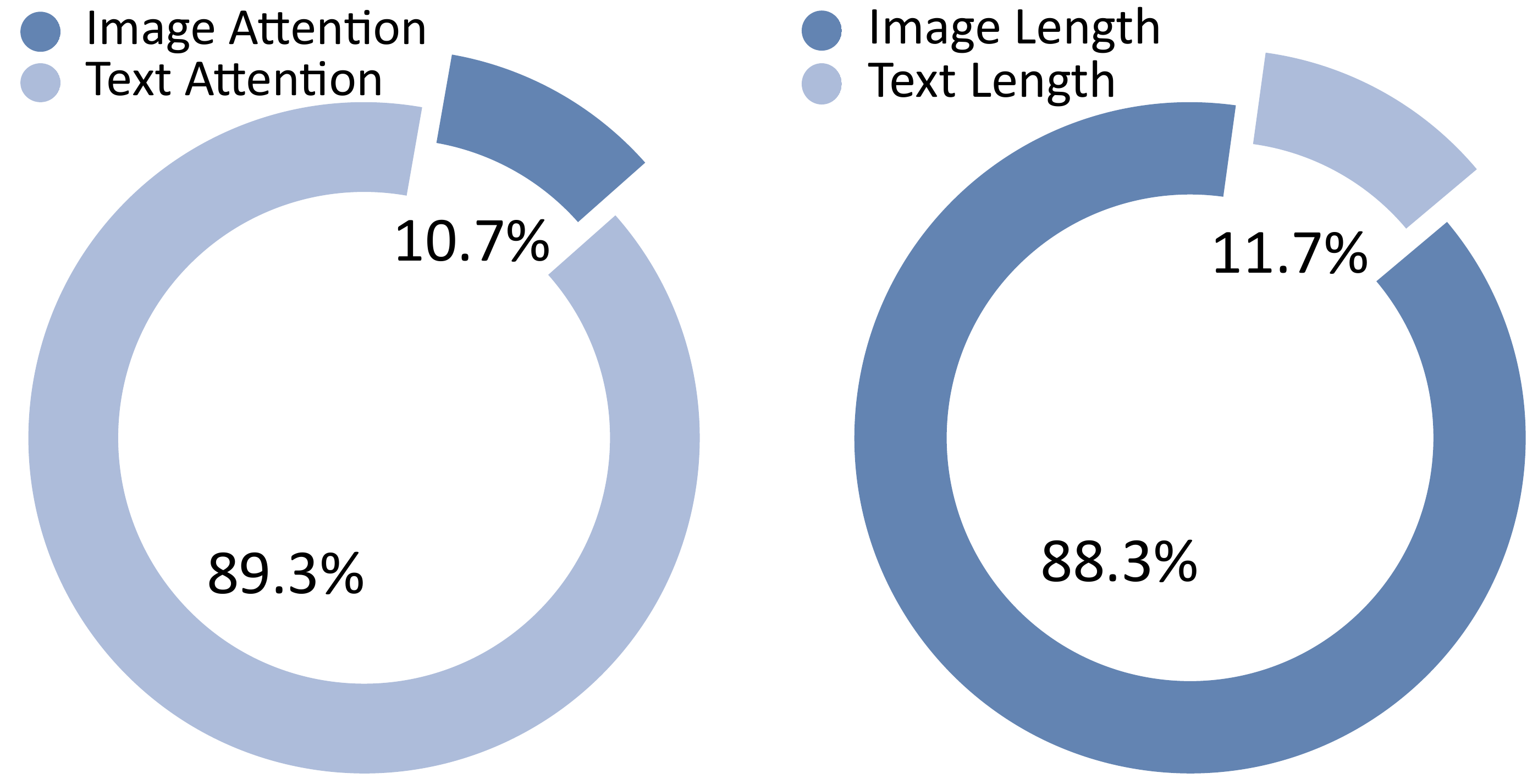}
     \vspace{-2mm}
    \caption{
    A striking imbalance between visual and textual attention: while image tokens take  approximately 90\% of the sequence length, they receive only about 10\% of the model's total attention on WhatsUp. This severe disparity in attention allocation suggests that VLMs fundamentally underutilize visual information.   
    }
    \vspace{-3mm}
    \label{fig:attn_sum}
\end{figure}

\label{sec:obs1}
We hypothesize that the failure on spatial reasoning stems from issues in the text-vision interactions, specifically due to insufficient attention being allocated to the image tokens. In this section, we systematically investigate the impact of the absolute values of attention logits on the image tokens during spatial reasoning.


\noindent\textbf{Experiment settings.} We select a widely-used spatial reasoning benchmark WhatsUp~\citep{kamath-etal-2023-whats}  since it contains both synthetic data and realistic data. 
The synthetic data (\textbf{Controlled\_Image}) features clean backgrounds with two objects, as shown in the upper example of Figure~\ref{fig:main-figure}. It comprises two subsets: \textbf{Controlled\_A}, with one large object (e.g., table) and one small object (e.g., cup), and \textbf{Controlled\_B} with two small objects (e.g., book and plate). We use \textbf{Cont\_A} and \textbf{Cont\_B} as abbreviations in the paper.
The realistic data, as shown in the lower image of Figure~\ref{fig:main-figure}, contains complex backgrounds with multiple objects, sourced from MS COCO~\citep{lin2014microsoft} and Visual Genome~\citep{krishna2017visual} (referred to as \textbf{COCO} and \textbf{VG} later, both datasets include captions involving either one object or two objects, referred to as \textbf{COCO\_one}, \textbf{COCO\_two}, \textbf{VG\_one}, and \textbf{VG\_two}, respectively). 
Compared with realistic images,
the synthetic images enables clearer observation of \vlm's inner workings with just two objects. Each image is paired with a ground truth caption describing the spatial relationship of two objects.

 We reformat the original $\langle$\textit{image}, \textit{caption}$\rangle$ setting into a generative question-answering setting $\langle$\textit{image}, \textit{question}, \textit{spatial\_label}$\rangle$, enabling evaluation of generative models like \vlm and tracing of internal states. Questions are generated using GPT-4~\citep{GPT4V}. Details in each subset with prompts are shown at Figure~\ref{tab:dataset}. This QA task asks about positions in a multiple-choice format, with possible answers like ``left", ``right", ``on"  and so on.

For evaluation, we use LLaVA-1.5~\citep{liu2024visual} throughout the analysis part (Section~\ref{sec:obs1} and~\ref{sec:obs2}). For metric, we apply \textbf{accuracy} of exact match as the primary metric. To maintain consistency in the label space across datasets, we use a four-option setting $\langle \textit{left}, \textit{right}, \textit{on}, \textit{under} \rangle$
 for the Controlled\_Image and COCO subsets, and a six-option setting $\langle \textit{left}, \textit{right}, \textit{on}, \textit{under}, \textit{behind}, \textit{front}\rangle$
 for the VG as it contains additional spatial annotations.

To better understand model performance, we analyze the \textbf{label distribution}, which provides insight into the representation of different spatial relationships within the dataset. 
In \textbf{Controlled\_Image}, labels are uniformly distributed across categories (e.g., equal number of samples for ``\textit{left}'', ``\textit{right}'', ``\textit{on}'', ``\textit{under}''). 
Another interesting feature of this dataset, which contributes to our choice to use it, is its contrastive setting. 
{Controlled\_Image} includes \textbf{pairs} of images with same objects in both ``\textit{left}'' and ``\textit{right}'' positions, and \textbf{sets} of same objects exhibiting ``\textit{left}'', ``\textit{right}'', ``\textit{on}'', and ``\textit{under}'' relationships.
It enables us further assess model performance using \textbf{pair accuracy} and \textbf{set accuracy}, requiring correct identification of all relationships within a pair or set, defined by \citet{kamath-etal-2023-whats}. It provides a comprehensive evaluation of spatial relationships. 

\subsection{\vlm allocate sparse attention to the image.
}
We analyze how output tokens attend to image tokens by extracting the attention logits across layers, and present the following key findings: 
The sum of attention scores to \textbf{the image tokens is significantly lower} than that to all the input text tokens, despite the considerably higher number of image tokens. In Figure~\ref{fig:attn_sum}, we focus on the attention scores from the first generated token and sum the attention allocated to the image tokens in average for all attention heads in all samples in WhatsUp. The results reveal that image tokens receive substantially less attention, with text tokens receiving approximately nine times more. Although the image sequence has a length of $576$, compared to the text sequence, which typically ranges from $30$ to $40$ tokens in our short question-answering setting, the model predominantly focuses on text when generating outputs. That is, image information is sparsely processed by the language model.


\subsection{Is the answer more accurate if the model sees the image more? Not really.}
\begin{figure}[t!]
  \centering
  \includegraphics[width=0.3\textwidth]{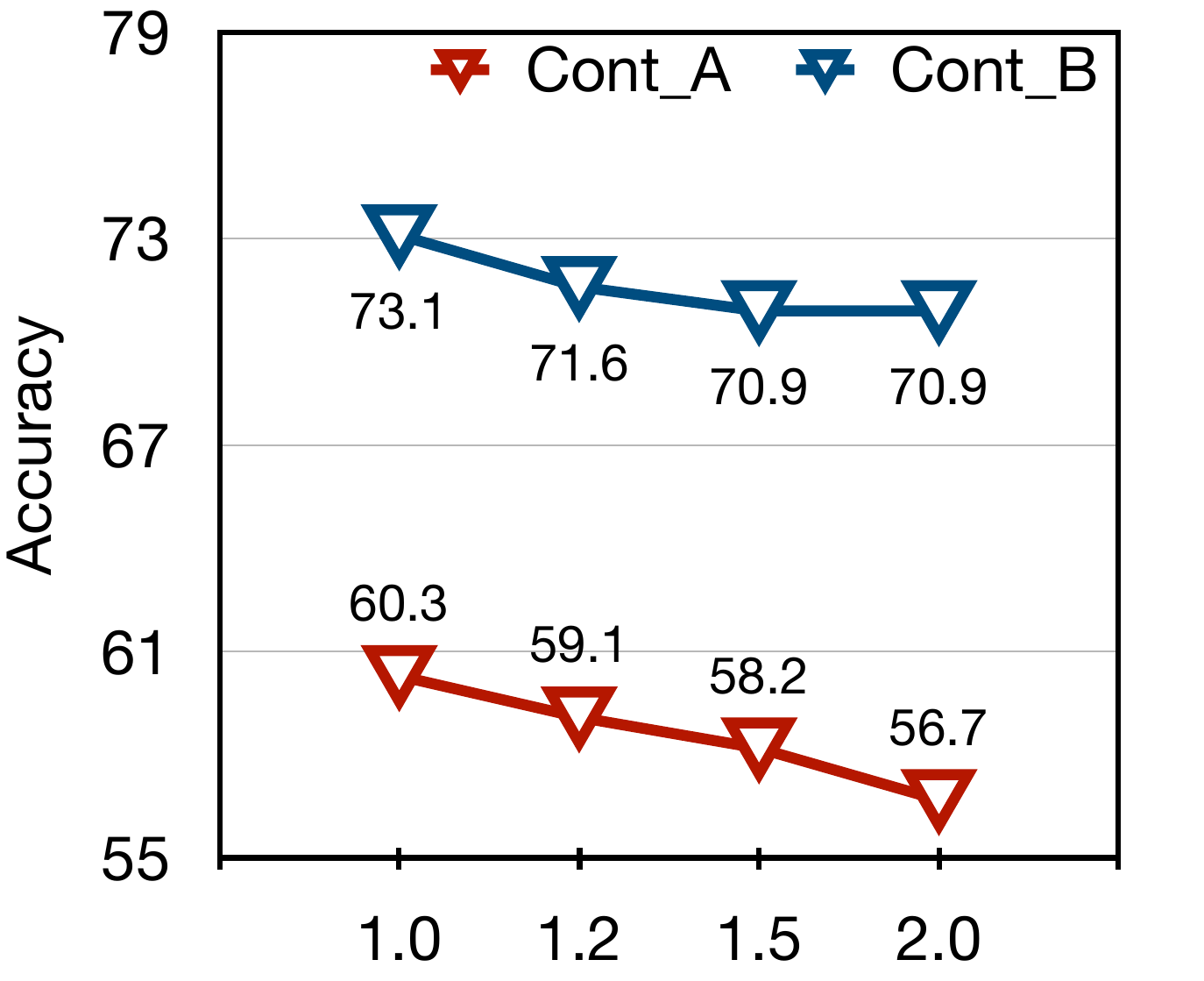}
  \vspace{-3mm}
  \caption{Accuracy of \textbf{adding} image attention in \textbf{logit space} (which corresponds to the multiplication operation in probability space; the x-axis of the figure represents the multiplication coefficient). \model, on the other hand, utilizes multiplication in logit space.}
  \label{fig:addattn_onall}
\end{figure}
Building on earlier observations, a natural question arises: since the attention scores assigned to the image are relatively low compared to those for text, could increasing attention to the image improve the factual accuracy of the model? To investigate this, 
we conduct an experiment by increasing the attention weights allocated to the entire image by the final answer, intervening by adding positive constants to the image attention logits across all patches
, as described by~\citet{zhang2023tell}. In Figure~\ref{fig:addattn_onall}, we observe that adding a constant weight uniformly across all image tokens does not improve performance on spatial reasoning tasks.
This observation motivates us to explore more intelligent ways to focus on key visual features.

\section{Visual Attention Distribution}
\label{sec:obs2}
\begin{figure*}[ht]
  \centering
  \begin{minipage}{0.244\textwidth}
    \centering
    \includegraphics[width=\textwidth]{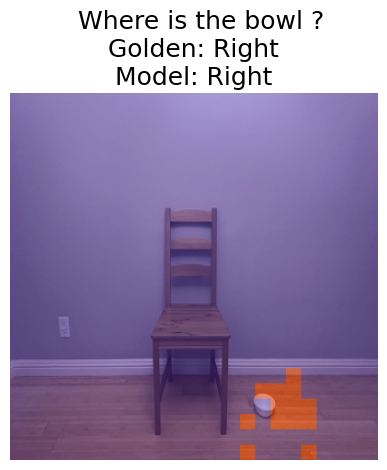}
  \end{minipage}\hfill
  \begin{minipage}{0.244\textwidth}
    \centering
    \includegraphics[width=\textwidth]{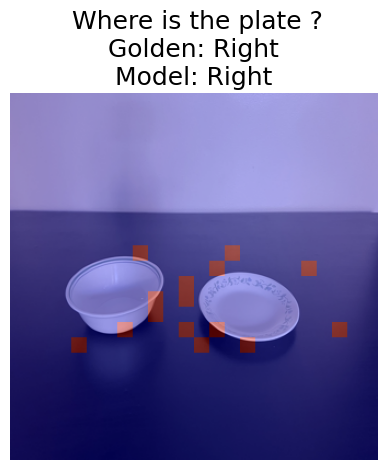}
  \end{minipage}
  \begin{minipage}{0.244\textwidth}
    \centering
    \includegraphics[width=\textwidth]{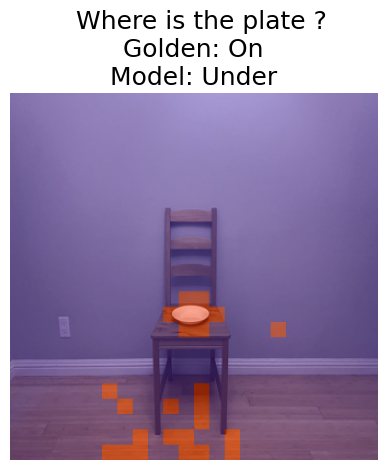}
  \end{minipage}
  \begin{minipage}{0.244\textwidth}
    \centering
\includegraphics[width=\textwidth]{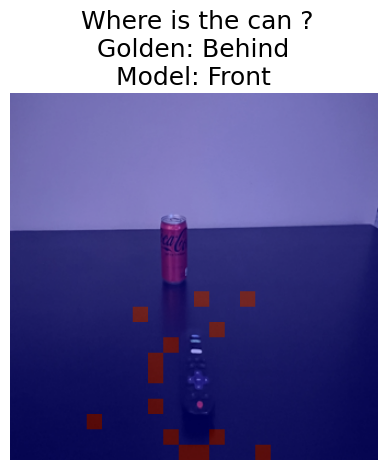}
  \end{minipage}
\caption{Attention visualization examples from the WhatsUp Dataset. The left two examples are answered correctly, while the right two are incorrect. For correctly answered questions, the attention scores are precisely focused on the core entities mentioned. In contrast, incorrect answers show attention scores distributed to irrelevant image regions. The visualizations use attention from the 17th layer, and the title in each image is an abbreviation of ``Where is A in relation to B". }
\label{fig:examples}
\end{figure*}
Our analysis of text-image attention imbalance has established that visual information is underutilized, yet simply increasing attention to image tokens fails to improve spatial reasoning. We hypothesize that the way attention is geometrically distributed across the image is the key factor. 
To investigate this hypothesis, we examine how attention patterns are distributed spatially by mapping the $576$ image tokens in LLaVA 1.5 to their $24 \times 24$ corresponding image patches. This mapping enables us to trace and visualize the geometric flow of attention, showing us not just where the model looks, but how its attention aligns with actual spatial relationships in the image.

\subsection{The model automatically focuses on the relevant entity when correctly answering questions.}
\label{sec:mislocate}

\begin{figure}[t!]
  \centering
  \begin{minipage}{0.5\textwidth}
    \centering
\includegraphics[width=\textwidth]{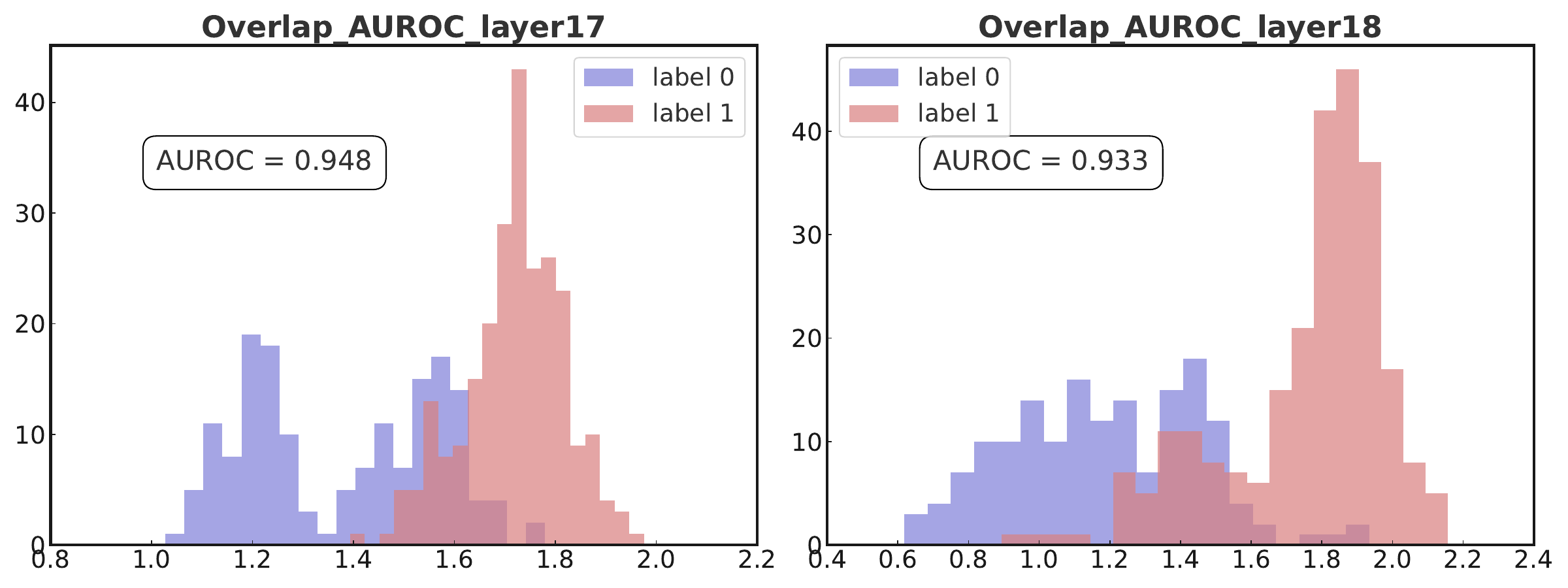}
  \end{minipage}
  \caption{The figure illustrates the AUROC of the overlap between YOLO annotations and the attention patterns for Cont\_A at the 17th and 18th layers. This high AUROC suggests that~\textbf{attention could be an effective metric for detecting answer correctness}. more AUROC results are shown at Appendix~\ref{appdix:more error case}.}
  \label{fig:overlap}
\end{figure}

To determine whether the model's factual inaccuracy is linked to incorrect attention behavior, we use YOLO~\citep{redmon2016you} to annotate the relevant entities in the images. We compare the overlap between these annotations with the model's attention distribution and evaluate whether the overlap can serve as a metric to predict the correctness of the model's answers.
We transform the bounding box coordinates obtained from YOLO annotations into image-sized index vectors, where each element is either 0 or 1. We then normalize these index vectors along with the attention logits of the image attended by the final answer of certain layers corresponding to the image patches.
We then calculate the cosine similarity of the two tensors for each sample, and analyze this metric on Cont\_A. As shown in Figure~\ref{fig:overlap}, the AUROC - a metric used to measure how well a model can distinguish between positive and negative classes.) is notably high in the middle layers, with results for all the layers at Figure~\ref{fig:middle}. Additional results on other subsets are presented in the Appendix~\ref{appdix: overlap}.

To investigate why the middle layers exhibit the most noticeable patterns, we analyze both attention scores and the overlap AUROC (Figure~\ref{fig:middle}) to track information flow and assess each layer's contribution. In the initial layers, attention to the input image is the highest compared to other layers. However, the AUROC in these layers is very low, suggesting that the model primarily captures a broad, global understanding rather than focusing on local details. In contrast, the middle layers play a more critical role in refining the model's understanding. As shown on the right side of Figure~\ref{fig:middle}, the overlap AUROC peaks in these layers, indicating that they are where the model begins to effectively ``process" image information. Additionally, the left side of Figure~\ref{fig:middle} shows a modest peak in attention logits, further supporting our hypothesis. 

\begin{figure}[t!]
    \centering
    \begin{minipage}{0.24\textwidth}
        \centering  \includegraphics[width=\textwidth]{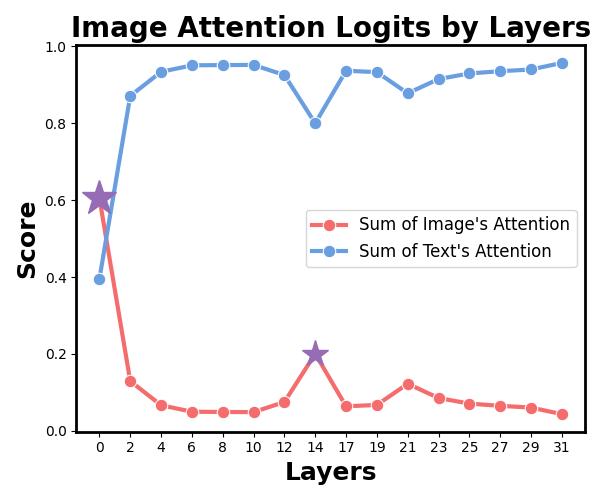}
    \end{minipage}\hfill
    \begin{minipage}{0.24\textwidth}
        \centering
        \includegraphics[width=\textwidth]{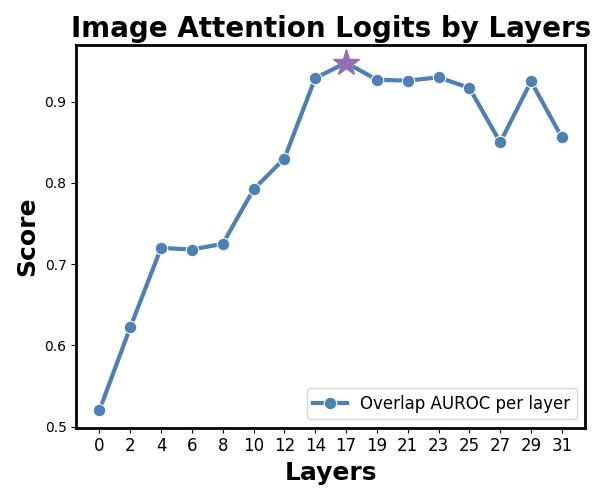}
    \end{minipage}
\caption{Left: Variance of image token attention across layers in Cont\_A, showing sparse attention on image, consistent with other subsets (Appendix~\ref{appdix:analysis}). Right: The Overlap's AUROC value across the layer of Cont\_A. We can infer from the two figures that the model ``sees" the image information in the early layers and ``processes'' this information in the intermediate layers. So for case study in our paper, we use the intermediate layers. }
\label{fig:middle}
\end{figure}

We further separately examine the attention patterns for correctly and incorrectly answered questions. Our observations reveal that hallucinations frequently occur due to two types of attention failures: (1) insufficient attention to the correct object, and (2) misplaced attention on irrelevant objects in the image. Figure~\ref{fig:examples} illustrates these findings. More examples are shown in Appendix~\ref{appdix: overlap}. In the two correct examples on the left, the attention scores are well-aligned with the referenced entities, with sufficient focus. On the other hand, the two incorrect examples on the right demonstrate how the model incorrectly assigns attention, effectively ``seeing'' the wrong parts of the image. While these examples highlight the qualitative differences in attention patterns, they do not provide a quantitative metric that aligns with our goal--developing a method to detect the reliability of internal states and enable intervention.  Therefore, the primary challenge lies in devising effective strategies to adjust the attention scores intelligently, given that we have no prior information about the scores until a single inference run generates the attention map.

\subsection{\textbf{\scal}: Temperature scaling to image attention distribution}
\label{sec:scalingvis}

Our observations from Section~\ref{sec:mislocate} reveal that the model often misallocates attention logits within images, leading to errors in spatial reasoning. 
To address this, we aim to enhance the model's ability to focus on key visual features, improving its capacity to accurately ground spatial relationships, especially in complex or ambiguous scenarios. To achieve this, we propose a straightforward yet effective method that dynamically adjusts the image attention by modifying the temperature of the attention distributions. By adjusting the temperature, we can make the distribution either sharper or smoother, effectively altering the attention distribution. For instance, if the attention pattern is mostly correct but lacks precision, we want to sharpens it. Conversely, if the attention pattern is fundamentally incorrect, we want to smooth it out, allowing us to explore alternative regions. 
This intervention targets the attention of the final input token (at the $n$-th position) to the image tokens.
\begin{equation}
    \mA_{n,j}^{(l,h)} = \begin{cases} \alpha \mA_{n,j}^{(l,h)} & \mathrm{if }\, j \in \mathcal{I} \\ \,\,\,\,\mA_{n,j}^{(l,h)} & \mathrm{otherwise}  \end{cases}
\end{equation}
where $\mathcal{I}$ represents the indices of all image tokens. In essence, we change the temperature of image attention distribution by multiplying a coefficient $\alpha$. From a physical perspective, multiplying by a coefficient greater than 1 encourages the model to focus more on the original distribution, while multiplying by a coefficient less than 1 makes the distribution more dispersed. In experiments, we uniformly apply this coefficient to all $H$ heads across all $L$ layers to avoid the need for extensive hyperparameter search. 

\begin{figure*}[!t]
  \centering
  \includegraphics[width=1.0\textwidth]{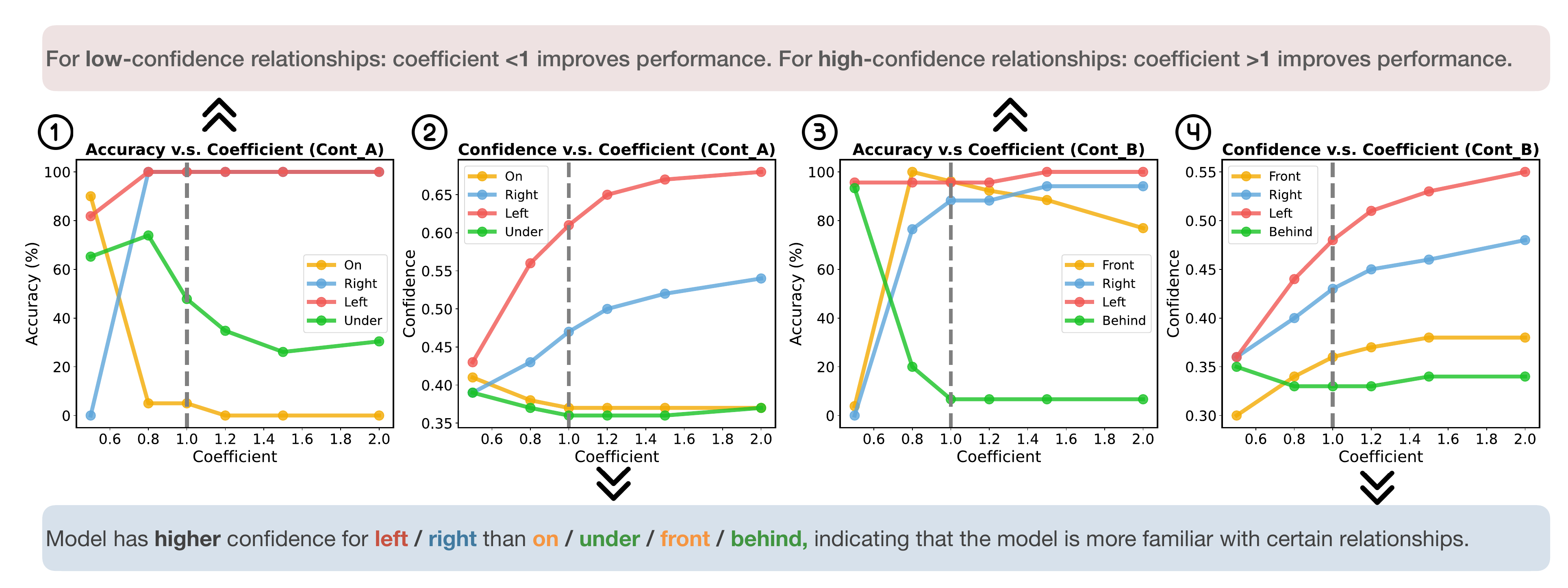}
  \caption{~\encircle[fill=white, text=black, draw=black, line width=1.5pt]{1} : Change in accuracy with the coefficient on Cont\_A.
~\encircle[fill=white, text=black, draw=black, line width=1.5pt]{2} : Change in confidence with the coefficient on Cont\_A.
~\encircle[fill=white, text=black, draw=black, line width=1.5pt]{3} : Change in accuracy with the coefficient on Cont\_B.
~\encircle[fill=white, text=black, draw=black, line width=1.5pt]{4} : Change in confidence with the coefficient on Cont\_B.
The baseline (greedy decode) corresponds to a coefficient of 1.  }
  \label{fig:conf}
\end{figure*}

\paragraph{Experiment Setting.}

\label{para: exp setting}
We select two widely-used benchmarks on evaluating the model's ability on spatial reasoning WhatsUp~\citep{kamath-etal-2023-whats} (introduced in Section~\ref{sec:background}), and  {VSR}~\citep{Liu2022VisualSR}, which contains contains $1223$ image-caption pairs with boolean labels. The original VSR is designed in $\langle$\textit{image}, \textit{caption}$\rangle$ format to evaluate encoder models without generation capabilities. To adapt it for our purposes, we utilize GPT-4o to generate questions for the VSR dataset. For evaluation, we report both accuracy and $F_1$ scores. A small validation set is allocated for each subset to optimize the temperature based on validation performance, and the final test is conducted on the test set. For both methods, the hyperparameter $\alpha$ is selected from $[0.5,0.8,1.2,1.5.2.0]$.  For baselines, we adopt DoLa~\citep{chuang2023dola} as a baseline, which employs the inner knowledge to calibrate the output logits by substracting the logits from intermediate layers. Additionally, we incorporate VCD~\citep{leng2024mitigating}, which is a contrastive decoding method to contrast the logits before and after eliminating the image.

\noindent\textbf{Results.} Our results for ScalingVis are presented in Table~\ref{tab:main_results} and Table~\ref{tab:vsr_results}. By controlling the distribution of attention weights, spatial reasoning performance improves significantly, with gains of up to 37.2 absolute points. An interesting pattern emerges: a temperature below one tends to enhance performance on synthetic data in most cases (3 out of 4), while a temperature above one benefits real image datasets across all cases. Table~\ref{tab:main_results} indicates that for \textbf{synthetic data}, smoothing the image attention logits improves performance. Conversely, for \textbf{real image datasets} (COCO and VG), the optimal temperature is consistently above one, demonstrating that a sharper attention distribution helps the language model discern relationships more effectively. Intuitively, we believe that for familiar datasets and spatial relationships, the model requires a sharper attention distribution, as it is generally correct but may not be sufficiently precise. In contrast, for unfamiliar datasets and spatial relationships, a smoother attention distribution is necessary to explore a broader context.

\begin{figure*}[t!]
  \centering
  \begin{minipage}{0.46\textwidth}
    \centering
    \includegraphics[width=\textwidth]{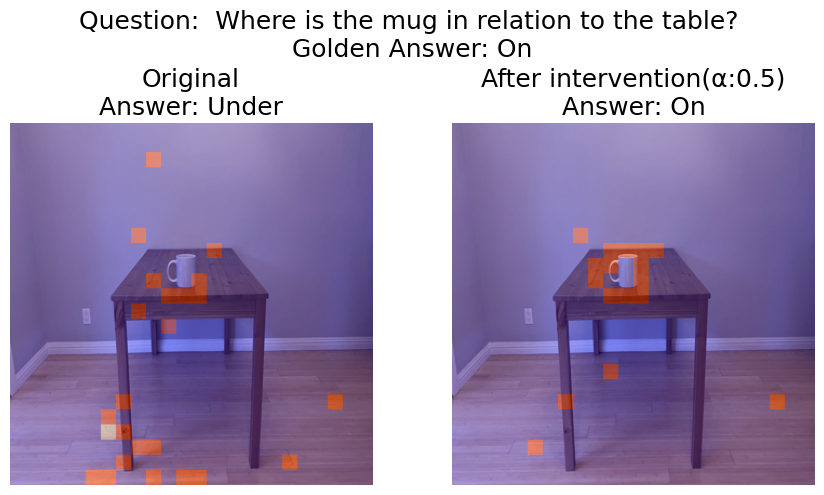}
  \end{minipage}\hfill
  \begin{minipage}{0.46\textwidth}
    \centering
    \includegraphics[width=\textwidth]{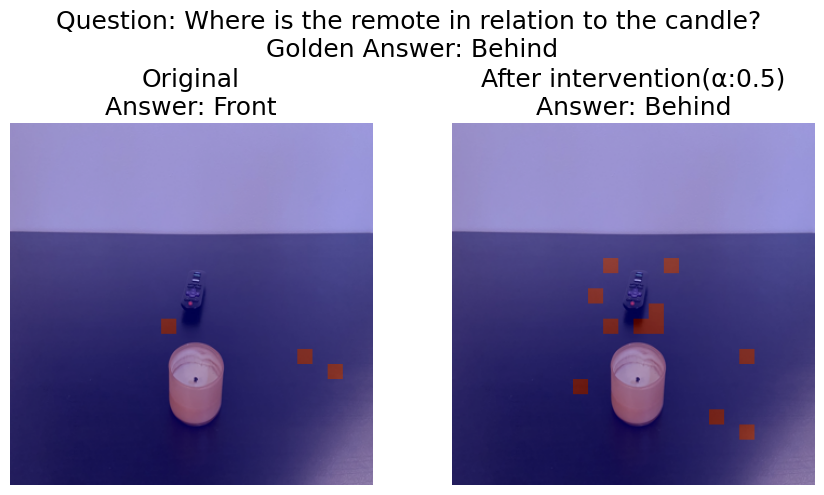}
    \vspace{-4mm}
  \end{minipage}
\caption{Attention scores on the patches before and after our intervention in the 17th layer for the images in \textbf{synthetic} datasets Cont\_A and Cont\_B. We employ a ``\textbf{smoothing}" intervention method to \textbf{expand the context length} of the model's focused area. From the figure, it is evident that the model's focused position undergoes significant changes after our intervention.
}
\label{fig:syn_case}
\end{figure*}

\begin{figure*}[t!]
  \centering
\begin{minipage}{0.46\textwidth}
    \centering
    \includegraphics[width=\textwidth]{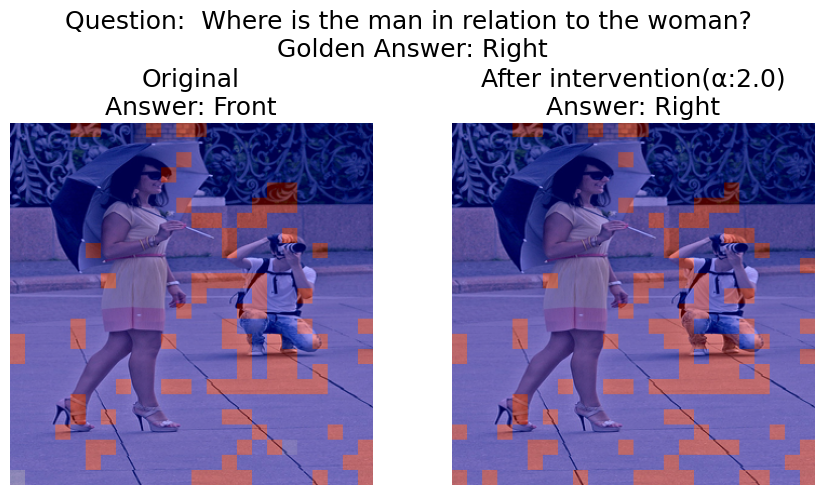}
  \end{minipage}\hfill
  \begin{minipage}{0.46\textwidth}
    \centering
    \includegraphics[width=\textwidth]{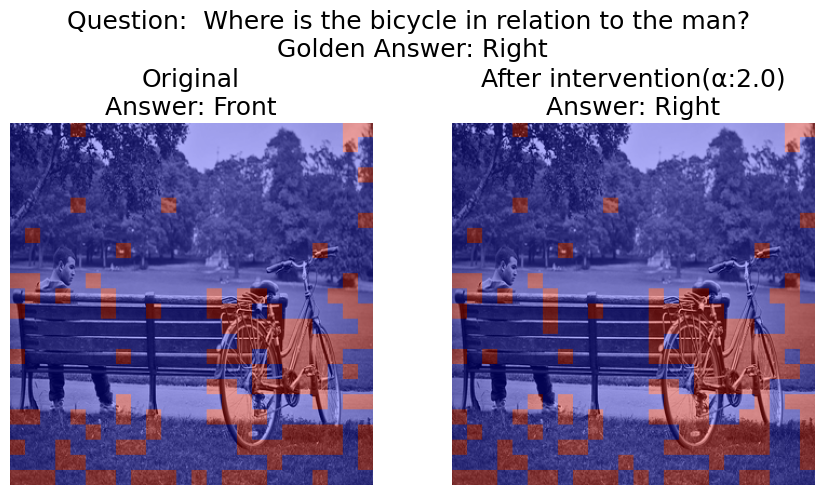}
  \end{minipage}
\caption{Attention scores on the patches before and after our intervention in the 17th layer for the images in \textbf{real} datasets COCO and VG. We utilize a ``\textbf{sharpening}" intervention method to enhance the original attention pattern. The highlighted areas remain largely consistent, with our method serving to reinforce the focus rather than significantly altering it.
 }
\label{fig:real_case}
\end{figure*}

\section{Adaptively Intervening the Attention Distribution by Model Confidence}

The findings from our study raise a key question: Since sharpening the distribution can sometimes improve performance while smoothing it is preferable in other cases, can we establish a metric to determine when to make these adjustments adaptively? In other words, can we establish a metric on whether we want to strengthen the original distribution or break it?
In this section, we explore how to assess when to trust image attention.
\subsection{Self-confidence reflects whether we can trust the model's image attention distribution.} \label{sec:confidence-intro}

We aim to determine whether an internal metric can indicate when the model’s attention pattern is trustworthy. Since we found that attention patterns often correlate with the correctness of the model’s generation as shown in~\textsection{
\ref{sec:mislocate}}, we reframe our goal as identifying a metric that helps assess the reliability of its outputs. Following previous findings of using generation logits as a measure of self-confidence to assess generation reliability~\citep{kadavath2022language, chen2024context}, we also adopt the logits of generated outputs to represent the model's self-confidence.
Specifically, we adjust the coefficient for all four spatial relationships in Cont\_A and Cont\_B, as these subsets provide a uniform distribution of different spatial relationships. We then examine how confidence and accuracy change in response and present the results in Figure~\ref{fig:conf}.
First, when the coefficient is 1.0 (baseline), the model exhibits significantly higher confidence for ``left'' and ``right'' relationships while showing notably lower confidence for other relationship types. This suggests that the model is more familiar with ``left'' and ``right'' relationships and less familiar with others. Furthermore, the accuracy of predicting ``left'' and ``right'' relationships is also higher, indicating the positive correlation between the model’s self-confidence and its correctness in predicting them.
Second, we observe distinct response patterns across different coefficients. Increasing the coefficient improves performance for ``left'' and ``right'' relationships, whereas decreasing it enhances performance for other relationships. This suggests that the model responds differently based on the relationship type, necessitating tailored intervention methods. These insights motivate us to propose an adaptive intervention strategy.

\subsection{\model}
Motivated by our observation that attention misallocation leads to spatial reasoning errors and that the model exhibits varying self-confidence and correctness across different spatial relationships, we propose \model: Confidence-aware temperature scaling, an extension of \scal: \noindent\textbf{Confidence-Based Attention Intervention.}

\begin{figure*}[!t]
  \centering
  \begin{subequations}
    \begin{minipage}{0.46\textwidth}
      \begin{equation}
      \label{eq:adapt-l}
        \mathbf{A}_{n,j}^{(l,h)} =
        \begin{cases} 
          \alpha_1 \mathbf{A}_{n,j}^{(l,h)} & \mathrm{if }\,j \in \mathcal{I} \\ \,\,\,\,\,\,\,
          \mathbf{A}_{n,j}^{(l,h)} & \mathrm{otherwise}
        \end{cases}, \mathrm{if }\,\, \mathcal{C} < \beta
      \end{equation}
    \end{minipage}%
    \hfill
    \begin{minipage}{0.46\textwidth}
      \begin{equation}
       \label{eq:adapt-s}
        \mathbf{A}_{n,j}^{(l,h)} =
        \begin{cases} 
          \alpha_2 \mathbf{A}_{n,j}^{(l,h)} & \mathrm{if }\, j \in \mathcal{I} \\ \,\,\,\,\,\,\,
          \mathbf{A}_{n,j}^{(l,h)} & \mathrm{otherwise}
        \end{cases}, \mathrm{if }\,\, \mathcal{C} > \beta
      \end{equation}
       
    \end{minipage}
  \end{subequations}
\end{figure*}

\definecolor{verylightgray}{gray}{0.9}
\begin{table*}[!t]
\centering
\large
\resizebox{1.0 \linewidth}{!}{
\begin{tabular}{lllllllllll}
\toprule
\multirow{2}{*}{\textbf{Model}} & \multicolumn{3}{c}{\textbf{Controlled\_A}} & \multicolumn{3}{c}{\textbf{Controlled\_B}} &  \textbf{COCO\_one} & \textbf{COCO\_two} & \textbf{VG\_one} & \textbf{VG\_two}\\
\cmidrule(lr){2-7} \cmidrule(lr){8-11}
  & \textbf{Acc} & \textbf{P Acc} &  \textbf{S Acc} & \textbf{Acc} & \textbf{P Acc} &  \textbf{S Acc} & \textbf{Acc} & \textbf{Acc} & \textbf{Acc}& \textbf{Acc} \\
\midrule
LLaVA-1.5  & 60.3 & 40.6 & 0.0 & 73.1 & 41.6 & 3.7 & 53.0 & 58.2 & 35.9 & 40.8 \\
+VCD & 61.5\up{1.2} & 39.4\downbad{1.2} & 0.0 & 73.4\up{0.3} & 42.2\up{0.6} & 3.7 & 53.3\up{0.3} & 58.2 & 35.8\downbad{0.1} & 42.5\up{1.7} \\
+DoLa & 61.2\up{0.9} & 41.6\up{1.0} & 0.0 & 73.4\up{0.3} & 42.2\up{0.6} & 3.7 & \textbf{53.7\up{0.7}} & 57.5\downbad{0.7} & 36.2\up{0.3} & 42.1\up{1.3} \\
+\cellcolor{verylightgray}\scal  & \cellcolor{verylightgray}64.5\up{4.2} & \cellcolor{verylightgray}40.6 &\cellcolor{verylightgray}0.0 & \cellcolor{verylightgray}75.2\up{2.1} &\cellcolor{verylightgray}44.6\up{3.0} &\cellcolor{verylightgray}9.8\up{6.1} &\cellcolor{verylightgray}53.6\up{0.6} &\cellcolor{verylightgray}59.4\up{1.2} &\cellcolor{verylightgray}42.7\up{6.8} & \cellcolor{verylightgray}48.1\up{7.3} \\
+\cellcolor{verylightgray}\model  & \cellcolor{verylightgray}\textbf{84.9\up{24.6}} & \cellcolor{verylightgray}\textbf{61.2\up{20.6}} & \cellcolor{verylightgray}\textbf{30.3\up{30.3}} & \cellcolor{verylightgray}\textbf{83.8\up{10.7}} & \cellcolor{verylightgray}\textbf{55.7\up{14.1}} & \cellcolor{verylightgray}\textbf{18.3\up{14.6}} &\cellcolor{verylightgray}53.6\up{0.6} & \cellcolor{verylightgray}\textbf{59.9\up{1.7}} & \cellcolor{verylightgray}\textbf{42.7\up{6.8}} & \cellcolor{verylightgray}\textbf{48.1\up{7.3}} \\
\midrule
LLaVA-1.6  & 48.2 & 37.6 & 0.0 & 63.0 & 39.1 & 3.7 & 59.7 & 41.8 & 31.6 & 7.3 \\
+VCD & 61.8\up{13.6} & 41.8\up{4.2} & 10.9\up{10.9} & 65.4\up{2.4} & 41.6\up{2.5} & 7.3\up{3.6} & 60.6\up{0.9} & 44.9\up{3.1} & 33.8\up{2.2} & 11.6\up{4.3} \\
+DoLa & 48.2 & 37.6 & 0.0 & 62.7\downbad{0.3} & 39.1 & 3.7 & 59.7 & 41.5\downbad{0.3} & 31.5\downbad{0.1} & 7.3 \\
+\cellcolor{verylightgray}\scal & \cellcolor{verylightgray}97.0\up{48.8} &\cellcolor{verylightgray}76.4\up{38.8} & \cellcolor{verylightgray}54.5\up{54.5} & \cellcolor{verylightgray}73.4\up{10.4} &\cellcolor{verylightgray}48.9\up{9.8} & \cellcolor{verylightgray}15.9\up{12.2} & \cellcolor{verylightgray}\textbf{63.1\up{3.4}} & \cellcolor{verylightgray}\textbf{47.7\up{5.9}} & \cellcolor{verylightgray}\textbf{38.2\up{6.6}} & \cellcolor{verylightgray}14.6\up{7.3} \\
+\cellcolor{verylightgray}\model & \cellcolor{verylightgray}\textbf{98.2\up{50.0}} & \cellcolor{verylightgray}\textbf{78.8\up{41.2}} & \cellcolor{verylightgray}\textbf{57.0\up{57.0}} &\cellcolor{verylightgray}\textbf{73.4\up{10.4}} & \cellcolor{verylightgray}\textbf{48.9\up{9.8}} & \cellcolor{verylightgray}\textbf{15.9\up{12.2}} & \cellcolor{verylightgray}\textbf{63.1\up{3.4}} & \cellcolor{verylightgray}\textbf{47.7\up{5.9}} & \cellcolor{verylightgray}35.2\up{3.6} & \cellcolor{verylightgray}\textbf{17.2\up{9.9}} 
  \\

\bottomrule
\end{tabular}}
\caption{
Results on Controlled A, Controlled B, COCO, and VG datasets. (Metrics in $\times 10^{-2}$). Best-performing method per model and dataset are highlighted in bold. P Acc and S Acc represents Pair Acc and Set Acc.
}
\label{tab:main_results}
\end{table*} 
\begin{table}[!t]
\centering
\renewcommand{\arraystretch}{1.0} 
\resizebox{1\linewidth}{!}{ 
\begin{tabular}{p{0.4\linewidth} p{0.3\linewidth} p{0.3\linewidth}}
\toprule
\textbf{Model} & \multicolumn{2}{c}{\textbf{VSR}} \\
\cmidrule(lr){2-3}
 & \textbf{Exact Match}  & \textbf{F1 Score} \\
\midrule
LLaVA-1.5  & 62.4 & 51.3 \\
+VCD & 62.4 & 50.6\downbad{0.7} \\
+DoLa & 62.8\up{0.4} & 53.2\up{1.9} \\
+\cellcolor{verylightgray}\scal  & \cellcolor{verylightgray}64.9\up{2.5} & \cellcolor{verylightgray}62.5\up{11.2} \\
+\cellcolor{verylightgray}\model  & \cellcolor{verylightgray}\textbf{65.0\up{2.6}} & \cellcolor{verylightgray}\textbf{62.5\up{11.2}} \\
\midrule
LLaVA-1.6  & 58.8 & 29.4 \\
+VCD & 58.8 & 29.4 \\
+DoLa & 59.3\up{0.5} & 31.2\up{1.8} \\
+\cellcolor{verylightgray}\scal  & \cellcolor{verylightgray}59.1\up{0.3} &\cellcolor{verylightgray}30.6\up{1.2} \\
+\cellcolor{verylightgray}\model  & \cellcolor{verylightgray}\textbf{62.7\up{3.9}} & \cellcolor{verylightgray}\textbf{39.3\up{9.9}} \\

\bottomrule
\end{tabular}}
\caption{
Results on the VSR dataset (Exact Match and F1) (Metrics in $\times 10^{-2}$). Best-performing method per model and dataset are highlighted in bold.
}
\label{tab:vsr_results}
\end{table}

Recall from Section~\ref{sec:confidence-intro} that we observe two distinct patterns in the model’s factuality behavior: (1) synthetic data presents more unfamiliar cases than the real data, and (2) \vlm could express uncertainty through confidence scores. These insights motivate using confidence scores as a metric for adaptive intervention in the model’s internal states. Our intuition is straightforward: when confidence is low, suggesting that the attention pattern may be unreliable, we smooth the attention distribution. This encourages the model to explore a broader range of image regions, increasing the likelihood of focusing on the correct patches. Conversely, when confidence is high and attention is dispersed across the image, we sharpen the distribution to concentrate on key objects more effectively. Specifically, we apply the targeted intervention to the attention of the last input token (at the \(n\)-th position) directed toward the image tokens as shown in Equation~\ref{eq:adapt-l},~\ref{eq:adapt-s}.
Overall, we use a large $\alpha > 1$ when the Confidence $\mathcal{C}$ is large, which sharpens the attention distribution, and the relevant objects are paid more attention to; we use a small $\alpha < 1$ when the confidence $\mathcal{C}$ is small, which mitigates the model's excessive concentration on certain image tokens and makes the overall attention distribution smoother across the image. 

\paragraph{Evaluation Settings.}

We employ the same setting with \scal as shown in \ref{para: exp setting}. For hyperparameter choice and robustness, we show in Appendix~\ref{subsec: adahyper}. 

\paragraph{Results}

Our main results are presented in Table~\ref{tab:main_results} and Table~\ref{tab:vsr_results}. By controlling the distribution of attention weights, we observe a ~\textbf{significant improvement in spatial reasoning ability, with gains of up to 50 absolute points}. In most cases, \model achieves the best performance, particularly for synthetic datasets like Cont\_A and Cont\_B, as shown in Table~\ref{tab:main_results}, where it significantly outperforms the generalized method \scal. These findings suggest that model performance varies considerably with the label distribution of the dataset, and smoothing the distribution (by applying a coefficient smaller than 1) enhances performance. For real-image datasets like COCO and VG, the adaptive method performs slightly better than the generalized approach, indicating the model's robustness across different label distributions. Example cases are shown in Figure~\ref{fig:syn_case} and Figure~\ref{fig:real_case}, illustrating the smooth effect and focus effect ($\alpha<1$ and $\alpha>1$), respectively. 
It is important to note that the LLaVA-series models are trained on the COCO dataset, which makes them highly confident and familiar with COCO and VG image types. Hence trusting the model’s self-belief and sharpening image attention improves performance. Notably, for datasets containing more unfamiliar images, the adaptive setting proves to be significantly more effective.

\section{Related Work}

The first line of related work focuses on the attention patterns in LMs. Some studies on attention patterns in LLMs reveal biased attention across context windows, such as ineffective use of the middle context~\citep{liu-etal-2024-lost} and initial token attention sinks~\citep{xiao2023streamingllm}. While some approaches use fine-tuning to overcome these biases~\citep{an2024make}, training-free methods like input-adaptive calibration~\citep{yu2024unveiling} and position-specific interventions~\citep{yu2024mitigate} offer efficient alternatives. PASTA~\citep{zhang2023tell}, a closely related method, emphasizes attention on selected segments for specific heads; we extend this to \vlm without manual segment specification or multiple validation runs. Our work is also related to failure analysis in \vlm, \vlm have been shown to hallucinate more in multi-object recognition tasks and rely on spurious correlations~\citep{chen2024multiobjecthallucinationvisionlanguagemodels}, with systematic visual limitations highlighted from a CLIP perspective~\citep{tong2024eyes}. Our work also connects to the decoding strategies for reducing hallucinations decoding strategies to mitigate hallucinations include contrastive decoding focusing on image regions~\citep{leng2024mitigating}, preference tuning through data augmentation~\citep{wang2024mdpoconditionalpreferenceoptimization}, and methods leveraging contrastive layers for enhanced knowledge extraction~\citep{chuang2023dola}, as well as activation-based optimal answer identification~\citep{chen2024context}.

\section{Conclusion and Future Work}
Our research uncovers the inner working mechanism of VLMs during spatial reasoning, which is a critical limitation in VLMs and constrains their practical utility when requiring geometric understanding of visual scenes.  
We identify critical insights through an in-depth study of attention behaviors across layers: 1) VLMs allocate surprisingly insufficient attention to image tokens; 2) the location of attention on image tokens is more crucial than quantity; and 3) generation confidence serves as a reliable indicator of its familiarity with the image and the correctness of its attention pattern. Based on these findings, we propose \model, a novel decoding method that dynamically adjusts attention distribution, significantly improving spatial reasoning performance. 
Future research could focus on further exploring the mechanism of VLMs on complicated geometric structure understanding, such as long-horizon spatial reasoning, and investigate other spatial reasoning bottlenecks. 

\section*{Impact Statement}

Our research into the spatial reasoning capabilities of Large Vision Language Models (VLMs) has significant implications across various domains of artificial intelligence and its real-world applications.
First and foremost, our findings highlight a critical limitation in current VLMs: while they excel at object recognition, they struggle with basic spatial relationships. This gap between recognition and spatial understanding has far-reaching consequences for the practical deployment of VLMs in scenarios requiring geometric comprehension of visual scenes. Industries such as robotics, autonomous navigation, and assistive technologies for the visually impaired are particularly affected. For instance, a robot that can identify objects but cannot understand their spatial relationships may struggle with tasks like picking and placing items or navigating complex environments. Similarly, autonomous vehicles might face challenges in interpreting traffic scenarios accurately, potentially compromising safety.

Our development of \model, a novel decoding method that dynamically adjusts attention distribution based on the model's confidence, represents a significant step forward. By enhancing VLMs' performance on spatial reasoning tasks, \model could unlock new possibilities in various fields. In healthcare, improved spatial reasoning could lead to more accurate interpretation of medical imaging, potentially improving diagnostic accuracy. In augmented reality applications, better spatial understanding could enable more immersive and interactive experiences. For assistive technologies, enhanced spatial reasoning could provide more accurate and useful descriptions of environments to visually impaired individuals, significantly improving their independence and quality of life.

Looking ahead, our work opens up new avenues for research in AI and cognitive science. The exploration of mechanism interpretability in VLMs, particularly for complex geometric structures and long-horizon spatial reasoning, could provide insights into how artificial systems process and understand spatial information. This could not only advance AI capabilities but also contribute to our understanding of human spatial cognition. Additionally, investigating the role of training data memorization in spatial reasoning bottlenecks could lead to more efficient and effective training methods for future AI models.


In conclusion, our research not only addresses a fundamental limitation in current VLMs but also paves the way for more versatile and capable AI systems. As we continue to advance VLMs' capabilities in visually-driven tasks requiring nuanced spatial understanding, we have the potential to significantly impact various sectors of society, from healthcare and assistive technologies to urban planning and environmental monitoring. The future research ahead in this field is both exciting and challenging, requiring ongoing collaboration between researchers, ethicists, and policymakers to ensure that these advancements benefit society as a whole.
\section*{Acknowledgments}
We thank Min Lin, Miao Xiong, Guangtao Zeng, Teng Xiao, Wei Liu, Chi Han, Chang Ma, Zhengxuan Wu, and Junlong Li for their valuable discussions and insights.


\bibliographystyle{icml2025}
\bibliography{main}

\clearpage
\appendix

\addcontentsline{toc}{section}{Appendix} 

\part{Appendix} 
\parttoc 

\section{Limitations} 
Firstly, our methods, \scal and \model, specifically address model-related spatial hallucinations and self-alignment issues but are not designed to handle errors outside the language model's capabilities, such as the CLIP failures discussed by~\citet{tong2024eyes}. Secondly, \model relies on distribution-based confidence to adaptively set the confidence threshold $\beta$, we also observe that the optimal $\alpha$ and $\beta$ is different across different distributions and prompts. This dependence on a validation set for tuning poses a limitation on its applicability.
\begin{figure*}[t!]
  \centering
  
  \begin{minipage}[b]{0.46\textwidth}
    \centering
    \includegraphics[width=1\textwidth]{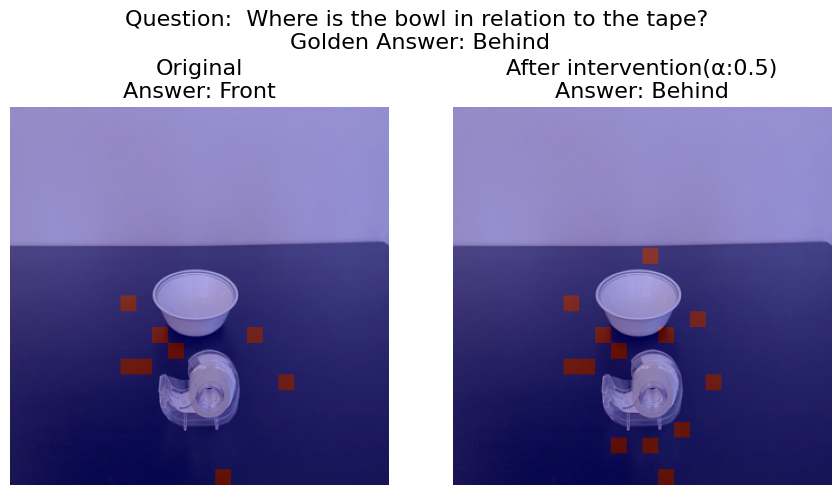}

  \end{minipage}
  \begin{minipage}[b]{0.46\textwidth}
    \centering
    \includegraphics[width=1\textwidth]{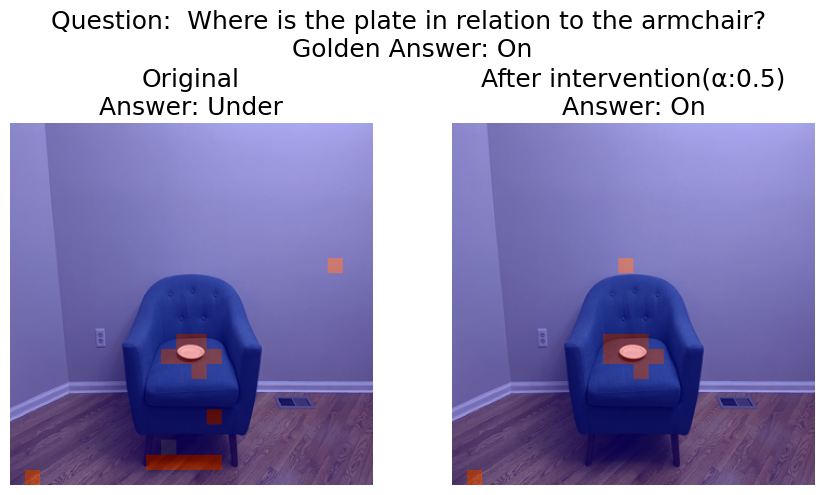}

  \end{minipage}
   \caption{Examples of our fixed case for $\alpha=0.5$. }
\label{fig:0.5fix}
\end{figure*}

\begin{figure*}[t!]
  \centering
  
  \begin{minipage}[b]{0.46\textwidth}
    \centering
    \includegraphics[width=1\textwidth]{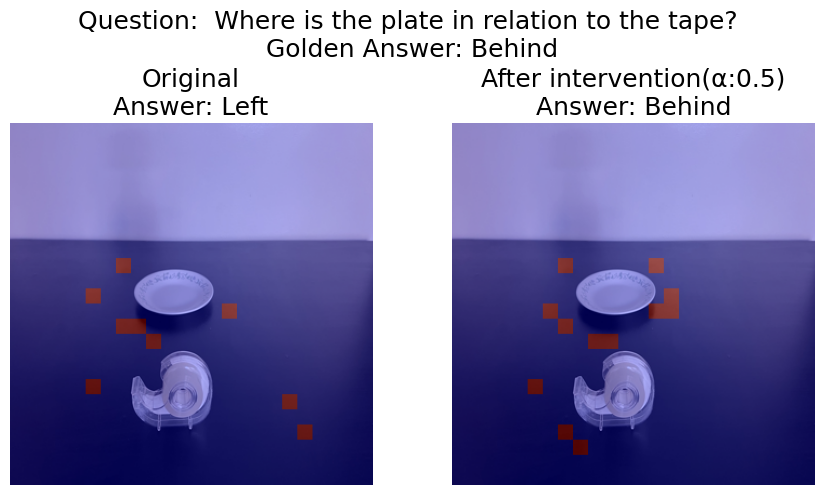}

  \end{minipage}
  \begin{minipage}[b]{0.46\textwidth}
    \centering
    \includegraphics[width=1\textwidth]{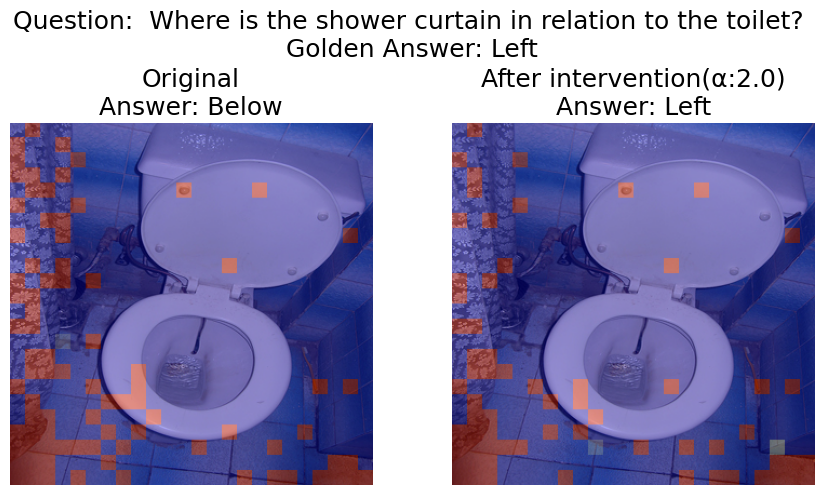}

  \end{minipage}
   \caption{Examples of our fixed case for $\alpha=2.0$. }
\label{fig:2.0fix}
\end{figure*}

\section{Additional Results on Qwen2-VL}
We further expand our intervention experiments and attention analysis on Qwen2-VL, the results on spatial reasoning benchmarks and on general benchmarks are shown in Table~\ref{tab:qwen2_attention_intervention} and Table~\ref{tab:qwen2_attention_intervention_pope} respectively. We intervene in the image attention distribution using our temperature-scaling method, showing consistent improvements below, particularly in challenging cases. For example, on VG\_two\_object, where the baseline performance is the lowest among all benchmarks, our method yields a significant improvement of 10+ absolute points. The gains observed on Qwen2-VL further demonstrate the generalizability of our approach. Experiments on more benchmarks shown in Table~\ref{tab:qwen2_attention_intervention_pope}, including POPE~\citep{li2023evaluating}, GQA~\citep{hudson2019gqa}, and TextVQA~\citep{singh2019towards}, which proves that attention intervention can maintain the performance on more general tasks without hurting the performance. Compared with spatial reasoning tasks, general QA tasks achieve relatively smaller improvement. A possible reason is that such tasks are less sensitive to the geometric structured distribution of image attention. For example, given a question like ``Is there a dog in this picture?”, the model only needs to detect the presence of an object, and is therefore less likely to suffer from misallocated attention across spatial regions.
\begin{table}[ht]
  \centering
  \begin{tabular}{@{}lcc@{}}
    \toprule
    \textbf{Benchmark} & \textbf{Qwen2-VL} & \textbf{+Attention Intervention} \\
    \midrule
    VSR             & 78.96 & 81.60\up{2.64} \\
    Coco\_one\_obj  & 76.64 & 78.03\up{1.39} \\
    Coco\_two\_obj  & 75.28 & 76.52\up{1.24} \\
    VG\_one\_obj    & 74.89 & 75.11\up{0.22} \\
    VG\_two\_obj    & 56.22 & 66.95\up{10.73} \\
    Controlled\_A   & 98.18 & 98.18\up{0.00} \\
    Controlled\_B   & 91.73 & 92.97\up{1.24} \\
    \bottomrule
  \end{tabular}
  \caption{Performance of Qwen2-VL before and after the attention intervention on spatial reasoning benchmarks.}
  \label{tab:qwen2_attention_intervention}
\end{table}

\begin{table}[ht]
  \centering
  \begin{tabular}{@{}lcc@{}}
    \toprule
    \textbf{Benchmark} & \textbf{Qwen2-VL} & \textbf{+Attention Intervention} \\
    \midrule
    POPE-Overall & 86.32 & 87.09\up{0.77} \\
    POPE-P       & 86.47 & 87.29\up{0.82} \\
    POPE-A       & 85.07 & 85.80\up{0.73} \\
    POPE-R       & 87.46 & 88.22\up{0.76} \\
    GQA          & 62.09 & 62.17\up{0.08} \\
    TextVQA      & 79.18 & 79.26\up{0.08} \\
    \bottomrule
  \end{tabular}
  \caption{Performance of Qwen2-VL before and after the attention intervention on additional general benchmarks.}
  \label{tab:qwen2_attention_intervention_pope}
\end{table}

We also conduct an attention analysis to verify whether our observation of the sparsity of image attention compared to textual tokens still holds. The results are shown in Table~\ref{tab:scalingvis_unc_acc}. Here we calculate the Qwen2-VL’s attention scores below (average attention logits for single text/image tokens respectively), which matches our previous claim that image receives much less attention than the text tokens is generally valid for VLMs. And we also conduct the uncertainty experiments to show our uncertainty conclusion is also valid. The results are shown in Table~\ref{tab:text_image_attention}. It shows that spatial relationships with lower confidence, such as ``Left" and ``Under" tend to exhibit lower accuracy when compared to those with higher confidence, like ``On" and ``Right". After the intervention with AdaptVis, certain spatial relationships like ``Left" show improved performance, accompanied by an increase in confidence. This demonstrates a pattern consistent with our observations for LLaVA, as depicted in Figure~\ref{fig:conf} in Section~\textsection{\ref{sec:obs2}}.
\begin{table}[ht]
  \centering
  \begin{tabular}{@{}lcc@{}}
    \toprule
    \textbf{Benchmark} & \textbf{Text attention} & \textbf{Image attention } \\
    \midrule
    Controlled\_A  & 1.57e-02 & 7.59e-05 \\
    Controlled\_B  & 1.58e-02 & 7.69e-05 \\
    Coco\_one\_obj & 1.77e-02 & 4.48e-04 \\
    Coco\_two\_obj & 1.65e-02 & 3.50e-04 \\
    VG\_one\_obj   & 1.54e-02 & 4.75e-04 \\
    VG\_two\_obj   & 1.42e-02 & 4.19e-04 \\
    \bottomrule
  \end{tabular}
  \caption{Average text-token and image-token attention scores across benchmarks.}
  \label{tab:text_image_attention}
\end{table}

\begin{table*}[ht]
  \centering
  \begin{tabular}{@{}lcccc@{}}
    \toprule
    \textbf{Benchmark} &
    \textbf{Qwen2‑VL Unc.} &
    \textbf{+ScalingVis Unc.} &
    \textbf{Qwen2‑VL Acc} &
    \textbf{+ScalingVis Acc} \\
    \midrule
    Left   & 0.4408 & 0.4474 & 70.11 & 74.71 \\
    On     & 0.5758 & 0.5668 & 100   & 100   \\
    Right  & 0.5982 & 0.5911 & 100   & 100   \\
    Under  & 0.5554 & 0.5418 & 98.82 & 98.82 \\
    \bottomrule
  \end{tabular}
  \caption{Uncertainty (\textit{Unc.}) and accuracy (\textit{Acc}) on spatial‑relation categories before and after applying ScalingVis.}
  \label{tab:scalingvis_unc_acc}
\end{table*}

\section{Related work}
\subsection{Attention Patterns in Language Models} Ongoing research has shown how large language models (LLMs) exhibit biased attention across different parts of the context window. \citet{liu-etal-2024-lost} find that LLMs fail to effectively utilize the information in the middle of a long context window. Meanwhile, \citet{xiao2023streamingllm} reveals an attention sink at the initial tokens of the input. Besides finetuning methods to overcome such biases~\citep{an2024make}, some training-free methods have been proposed with the benefit of their efficiency. \citet{yu2024unveiling} proposes to use input-adaptive calibration to adjust the attention scores, while \citet{yu2024mitigate} intervenes in position-specific hidden dimensions to alleviate the lost-in-the-middle phenomenon. A closely related work to ours is PASTA~\citep{zhang2023tell}, which emphasizes the attention scores of specific text segments for selected attention heads. We further develop this motivation on vision language models. Moreover, our method does not require a manual specification of the emphasized segment or multiple validation runs to identify effective attention heads. 

\subsection{Failure Analysis of Vision-Language Models}  
Our work relates to research on hallucination detection in \vlm. ~\citet{chen2024multiobjecthallucinationvisionlanguagemodels} examine multi-object recognition tasks, observing that \vlm exhibit more hallucinations when dealing with multiple objects compared to single-object scenarios. They also note a similar phenomenon to our findings: the distribution of tested object classes impacts hallucination behaviors, suggesting that \vlm may rely on shortcuts and spurious correlations. Additionally, \citet{tong2024eyes} analyze VLM failures from a CLIP perspective, highlighting that the visual capabilities of recent \vlm still face systematic shortcomings, partly due to CLIP’s limitations in specific cases.

\subsection{Decoding Strategies for Reducing Hallucinations}  
This work is also connected to various decoding and tuning strategies aimed at mitigating hallucinations in \vlm. ~\citet{leng2024mitigating} introduce a contrastive decoding method that emphasizes certain image regions. ~\citet{wang2024mdpoconditionalpreferenceoptimization} propose a data-augmentation approach to create image-intensive datasets, followed by preference tuning on this enhanced data. Furthermore, knowledge extraction techniques such as the method proposed by ~\citet{chuang2023dola} improve decoding by leveraging contrastive layers for better knowledge extraction. Similarly, Activation Decoding~\citep{chen2024context} identifies optimal answers as those with the highest activation values within the context.

\section{Case Study}
\label{appdix:more error case}
\begin{figure*}[t!]
  \centering
  \begin{minipage}{0.22\textwidth}
    \centering
    \includegraphics[width=1\textwidth]{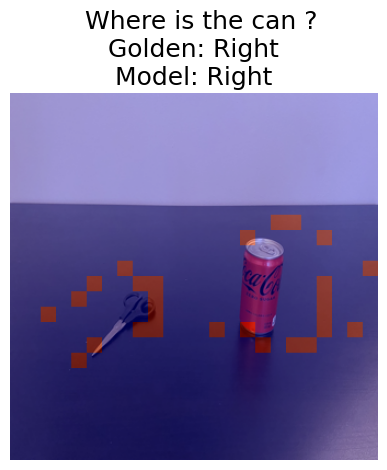}
  \end{minipage}
  \begin{minipage}{0.22\textwidth}
    \centering
    \includegraphics[width=1\textwidth]{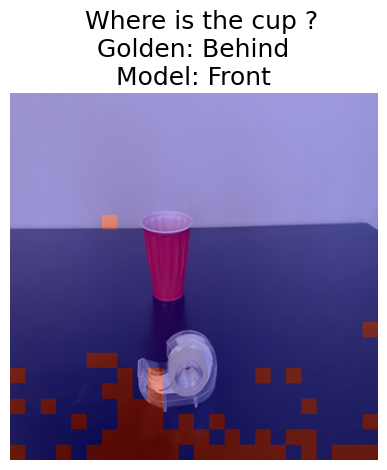}
  \end{minipage}
  \begin{minipage}{0.22\textwidth}
    \centering
    \includegraphics[width=1\textwidth]{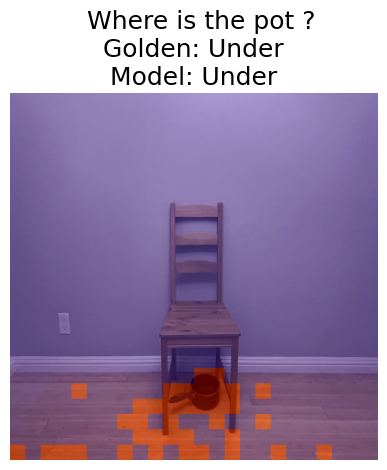}
  \end{minipage}
  \begin{minipage}{0.22\textwidth}
    \centering
    \includegraphics[width=1\textwidth]{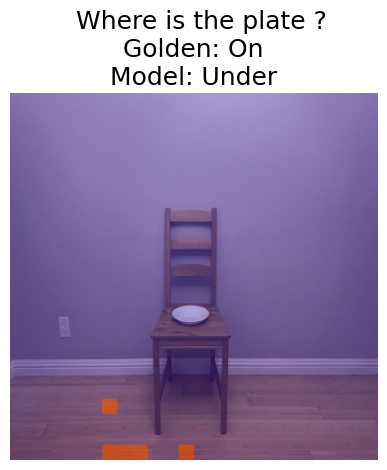}

  \end{minipage}
   \caption{Examples of image attentions. Visualization here is adopted from 14th layers' attention scores.}
    \label{fig:exampleattn}
\end{figure*}

We show more case we could fix in Figure~\ref{fig:0.5fix} and Figure~\ref{fig:2.0fix}.
We show more attention examples at Figure~\ref{fig:exampleattn}.

\section{Further Analysis}

\subsection{Other metrics to distinguish the distributions}
\label{appdix:other metrics}
\begin{figure*}[t!]
  \centering
  
  \begin{minipage}{0.38\textwidth}
    \centering
    \includegraphics[width=1\textwidth]{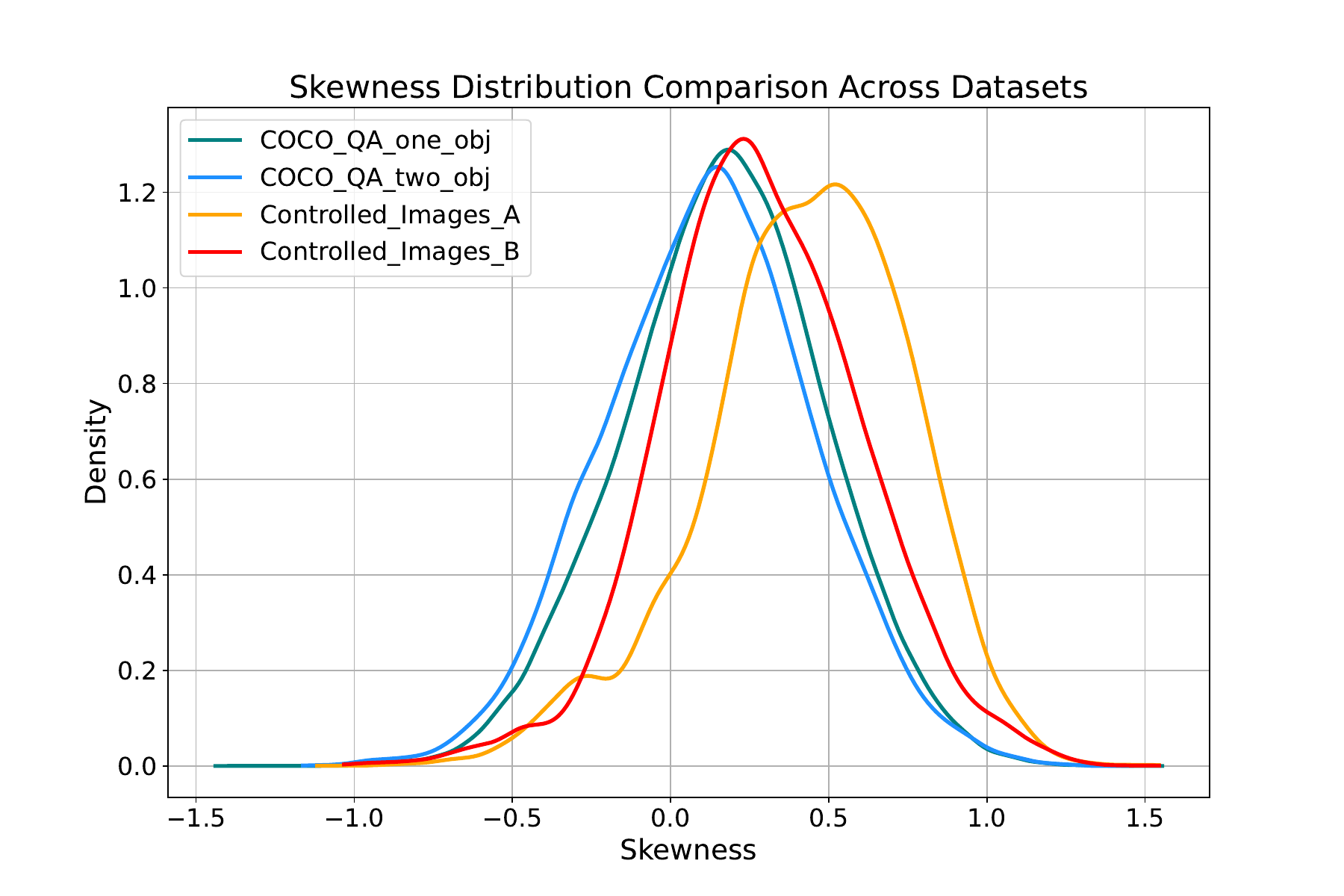}
  \end{minipage}
    \label{fig:skewness}
  \begin{minipage}{0.38\textwidth}
    \centering
    \includegraphics[width=1\textwidth]{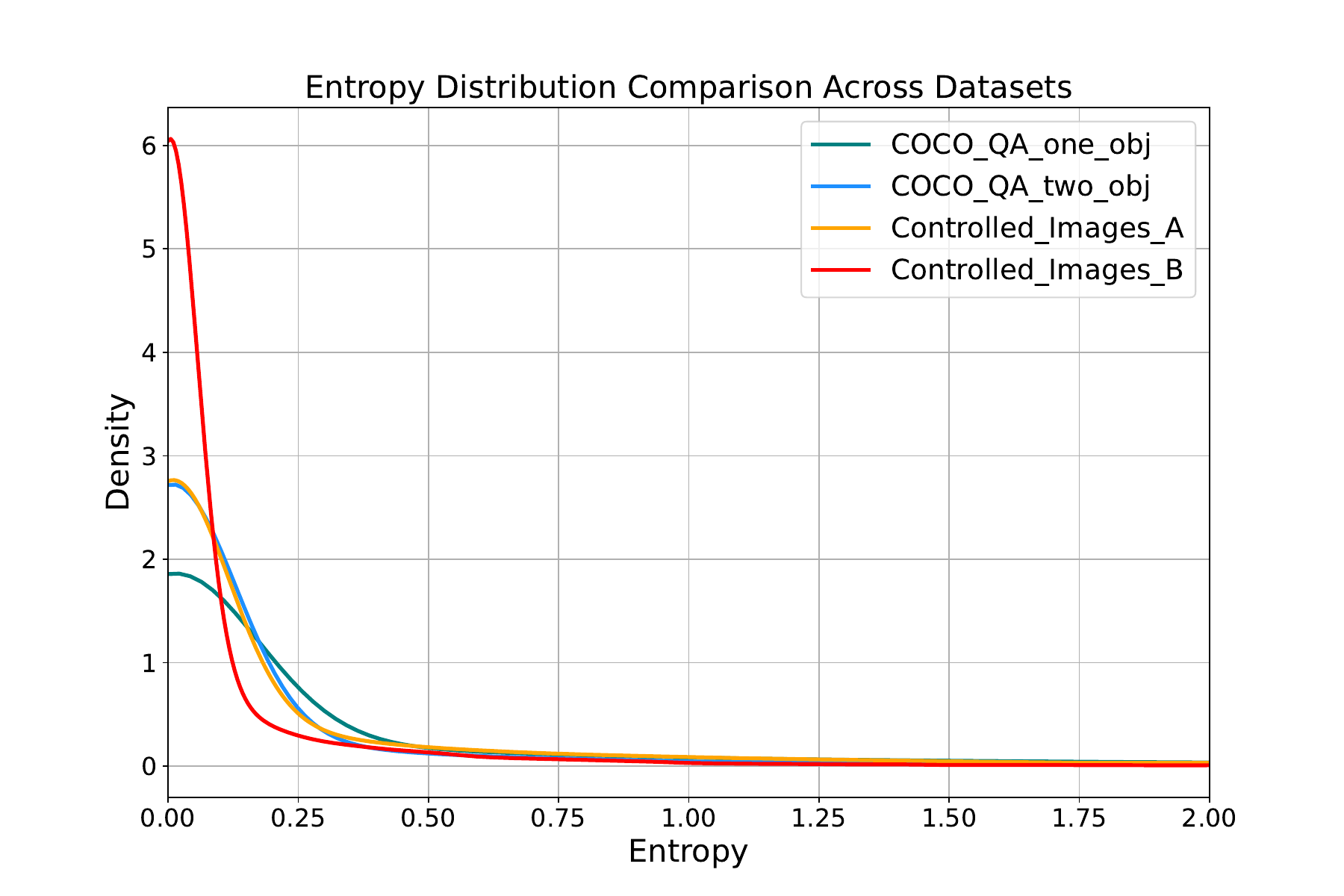}

  \end{minipage}
   \caption{The skewness and entropy distribution comparison between different subsets. Here we use Controlled\_Images and COCO datasets due to all of them are in four option label space, which enables us to eliminate the influence of prompts. }
    \label{fig:entropy}
\end{figure*}

We further explored additional metrics to distinguish the distributions and uncover their underlying characteristics during our preliminary experiments. Specifically, we examined entropy and skewness. Entropy was selected based on the hypothesis that parameter differences may stem from the familiarity of attention patterns in real images, which are generally correct, versus synthetic images, where these patterns tend to be incorrect. We posit that the model can express "familiarity" through certain metrics derived from the attention scores. For example, we hypothesize that the entropy of attention will be lower when the model encounters familiar cases.

\begin{equation}
E\left( \mathcal{A}_{n,j}^{(l,h)} \right) = -\sum_{j=1}^{t} \tilde{P}\left( \mathcal{A}_{n,j}^{(l,h)} \right) \log \tilde{P}\left(  \mathcal{A}_{n,j}^{(l,h)} \right)
\label{eq:entropy}
\end{equation}
In Equation~\ref{eq:entropy},\(\mathcal{A}_{n,j}^{(l,h)}\) denotes the attention scores assigned by the \(h\)-th head in the \(l\)-th layer to the \(j\)-th token in sequence \(n\). The summation runs over \(j = 1\) to \(t\), where \(t\) is the total number of tokens considered for this attention distribution. \(\tilde{P}\left( \mathcal{A}_{n,j}^{(l,h)} \right)\) is the normalized probability distribution of these attention scores. This entropy measures the uncertainty or spread of the attention distribution across tokens.

Our experimental results in Figure~\ref{fig:entropy} indicate that the attention distribution is heavily influenced by image features. Notably, the attention distribution is more concentrated in synthetic datasets than in real images. We attribute this to the fact that synthetic images tend to contain fewer objects, resulting in a sharper attention distribution. However, this concentration does not provide a reliable metric for measuring familiarity. Another possible metric is skewness. Another possible metric is skewness, which captures the asymmetry of the attention distribution. A high skewness suggests that the attention is predominantly focused on a few positions, while a low skewness indicates a more balanced spread across multiple regions. By examining skewness, we aim to identify whether the attention is being disproportionately allocated to particular image features, which could provide additional insights into how familiarity is expressed through attention patterns.
We could see from Figure~\ref{fig:entropy} that Synthetic datasets show a higher skewness. However, it also related with the object distribution, which is not the real factor behinds the difference.

\begin{equation}
S\left( \mathcal{A}_{n,j}^{(l,h)} \right) = \frac{\sum_{j=1}^t \left( j - \mu_{\mathcal{A}} \right)^3 \tilde{P}_j}{\sigma_{\mathcal{A}}^3}
\end{equation}
 The summation runs over \(j = 1\) to \(t\), where \(t\) is the number of tokens considered in the attention distribution. \(\tilde{P}_j\) denotes the probability assigned to the \(j\)-th token in the normalized attention distribution. The term \(\mu_{\mathcal{A}}\) is the mean of the distribution, and \(\sigma_{\mathcal{A}}\) is its standard deviation. The skewness is calculated as the normalized third central moment, which measures the asymmetry of the attention distribution: a positive value indicates a distribution skewed to the right, while a negative value indicates a skew to the left.

\subsection{Label statistics in WhatsUp}
\begin{table} 
\centering
\small
\resizebox{1.0 \linewidth}{!}{
\begin{tabular}{lcccccc}
\toprule
\multirow{2}{*}{\textbf{Dataset}} & \multicolumn{6}{c}{\textbf{Relationship Types}} \\
\cmidrule(lr){2-7}

 & \textbf{Right} & \textbf{Left} & \textbf{On} & \textbf{Under} & \textbf{Behind} & \textbf{Front} \\
\midrule
Controlled\_A & 92  & 92 & 130     & 92  & 0   & 0 \\
Controlled\_B & 102  & 102 & 0     & 0     & 102   & 102 \\
VG\_one   & 376  & 392  & 192   & 198   & 2   & 0 \\
VG\_two   & 137  & 127  & 3     & 0     & 5   & 19 \\
COCO\_one & 564  & 576  & 363   & 744   & 0   & 0 \\
COCO\_two & 129  & 150  & 86    & 75    & 0   & 0 \\

\bottomrule
\end{tabular}}
\caption{Gold Answer Frequency by Spatial Relation in WhatsUp dataset.}
\label{tab:gold_answer_freq_extra}
\end{table}

We present the golden labels' distribution in Table~\ref{tab:gold_answer_freq_extra}. We can see that the label space in synthetic datasets is balanced while the real image datasets are imbalanced with more ``left" and ``right" and fewer other relationships.

\begin{figure*}[t!]
  \centering
  \includegraphics[width=1.0\textwidth]{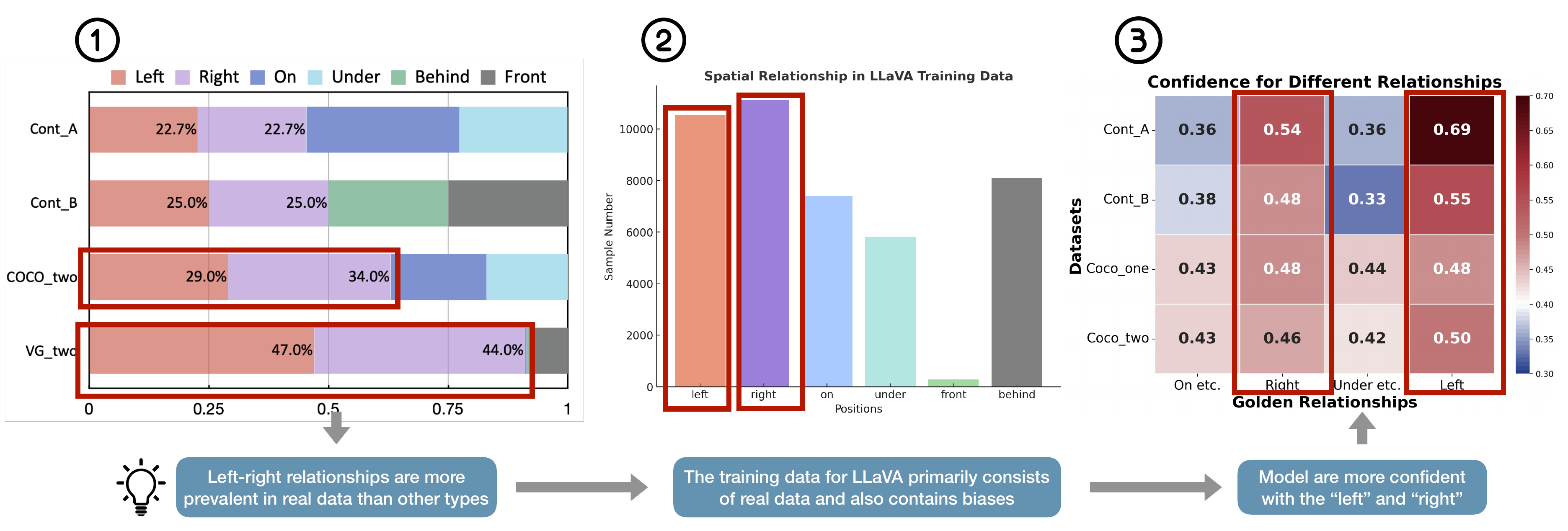}
  \caption{~\encircle[fill=white, text=black, draw=black, line width=1.5pt]{ 1}: Label distributions across subsets capturing relationships between object pairs in WhatsUp. For~\textbf{real datasets} like COCO and VG, a severe~\textbf{label imbalance} is evident, with ~\textbf{``left"} and~\textbf{``right"} being more common than other relationships.~\encircle[fill=white, text=black, draw=black, line width=1.5pt]{ 2} show the statistics in LLaVA training data. We could see imbalance problem still holds. ~\encircle[fill=white, text=black, draw=black, line width=1.5pt]{ 3}: The model's average confidence in different golden spatial relationships within these four-option subsets in WhatsUp. ``On etc'' includes ``\textit{on}, \textit{above}, \textit{top}'', while ``Under etc'' includes ``\textit{under}, \textit{below}, \textit{bottom}''. The red heatmap box highlights instances where the model is confident in its generation, while the blue box indicates the opposite. We can observe that ~\textbf{for familiar relationships}, such as ``left'' and ``right'' (highlighted with two red boxes),~\textbf{the model shows higher confidence.} }
  \label{fig:label_dist}
\end{figure*}

\paragraph{Label Distribution and the familarity}
To determine whether to focus more on the image or the text, we begin by analyzing existing datasets to identify potential patterns. Our investigation starts with an examination of label distributions across different subsets of the WhatsUp dataset. As shown in Figure~\ref{fig:label_dist}, there is a clear label imbalance in the real-image datasets including COCO\_two and VG\_two, i.e., only a small portion of the samples have the relation of ``\textit{behind}'' and ``\textit{front}''. This may be attributed to annotation bias, where left and right are easier to distinguish than other relationships. In contrast, the synthetic datasets, Cont\_A and Cont\_B, which are carefully curated, display more balanced label distributions. Additionally, we evaluate the model's confidence scores across the ground-truth spatial relations by following the approach of ~\citet{kadavath2022language}, where the probability of the generated output is used to compute confidence. Figure~\ref{fig:label_dist}  reveals that \textbf{the model is unconfident with specific spatial relationships}, such as ``\textit{on}'' and ``\textit{under}'' where it shows a colder color at Figure~\ref{fig:label_dist}, while \textbf{demonstrating higher confidence in recognizing more common relationships} where it shows a warmer color. In Figure~\ref{fig:conf}, we observe that the model performs better on confident relationships. For example, in the first figure on the left, when the coefficient is set to 1 (the baseline), the ground-truth samples labeled as \textit{``left"} and \textit{``right"} exhibit significantly better performance compared to those labeled as \textit{``on"} and \textit{``under"}. This observation aligns with our intuition that model tends to be more confident when it performs well and less confident when it struggles. This finding is also consistent with previous work showing that models can convey their uncertainty through confidence scores ~\citep{kadavath2022language, xiongcan}. Motivated by this, we propose using confidence as a metric to gauge the model’s familiarity with spatial relationships in images.

\paragraph{Statistics of LLaVA training data}
To categorize six different spatial relationships, we check whether specific phrases appear in the ``llava\_v1\_5\_mix665k.json", which is the training data of LLaVA and increment corresponding counters. For the \textbf{left} relationship, we look for phrases such as ``left side,'' ``left of,'' ``to the left,'' and ``on the left.'' Similarly, for the \textbf{right} relationship, we detect phrases like ``right side,'' ``right of,'' ``to the right,'' and ``on the right.'' The \textbf{on} relationship is identified using phrases like ``are on the,'' ``is on the,'' and ``located on,'' ensuring they do not co-occur with ``on the left'' or ``on the right.'' To detect the \textbf{under} relationship, we check for phrases such as ``under the,'' ``beneath the,'' or ``below the.'' For the \textbf{front} relationship, we look for phrases like ``are in front of,'' ``is in front of,'' and ``locate in front of.'' Finally, the \textbf{behind} relationship is identified by the phrase ``behind the.''. The results are shown in Figure~\ref{fig:label_dist}.

\subsection{Hyperparameter choice and robustness}
\label{subsec: adahyper}
\paragraph{For \model} We use the same hyperparameters for \scal as in Section~\ref{sec:scalingvis}. For \model, we optimize $\alpha_1$, $\alpha_2$, and $\beta$ using the validation set (randomly sampled $20\%$ from each subset) from each distribution. We also show our robustness for hyperparameters in Appendix. Notably, we find that model performance is robust across a range of these hyperparameters, generalizing effectively to other subsets within the same distribution (as demonstrated in Table~\ref{tab:ood}). We maintain consistency by using the same range of $\alpha$ values for both methods. For $\beta$, we adjust per dataset: for WhatsUP, we select values from $[0.3, 0.65]$ for LLaVA-1.6 and $[0.2, 0.55]$ for LLaVA-1.5 with a grid size of $0.05$ (this higher range is due to LLaVA-1.6 generally exhibiting higher confidence than LLaVA-1.5); for VSR, we take the mean value of the average confidence scores corresponding to the two labels.

\paragraph{Out-of-Domain Test of Hyperparameters.}
We evaluated the generalizability of common hyperparameters across datasets. To this end, we applied the same set of four-option prompts to Controlled\_Images and COCO subsets. Results in Table \ref{tab:ood} indicate that \model consistently performs well across all subsets, confirming its generalizability. 
\begin{table}[h]
\centering
\Large
\resizebox{0.5\textwidth}{!}{
\begin{tabular}{llllllllll}
\toprule
\multirow{2}{*}{\textbf{Model}} & \multicolumn{3}{c}{\textbf{Cont\_A}} & \multicolumn{3}{c}{\textbf{Cont\_B}} & \textbf{Coco\_one} & \textbf{Coco\_two} \\
\cmidrule(lr){2-9}
  & \textbf{Acc} & \textbf{P-Acc} & \textbf{S-Acc} & \textbf{Acc} & \textbf{P-Acc} & \textbf{S-Acc} & \textbf{Acc} & \textbf{Acc} \\
\midrule
LLaVA & 60.3 & 40.6 & 0.0 & 73.1 & 41.6 & 3.7 & 53.0 & 58.2 \\
+Ours & 60.3 & 41.8\up{1.2} & 2.4\up{2.4} & 76.5\up{3.4} & 48.3\up{6.7} & 13.5\up{9.8} & 53.6\up{0.6} & 59.4\up{1.2} \\
\midrule
Best $\alpha$ & \multicolumn{8}{c}{\cellcolor{ood} $\alpha_1=0.5 \quad \alpha_2=1.2 \quad \beta=0.3$} \\
\bottomrule
\end{tabular}}
\vspace{-2mm}
\caption{
OOD test results on WhatsUp (Metrics in $\times 10^{-2}$) for LLaVA1.5. Arrows show growth over baseline. P-Acc and S-Acc are Pair and Set Accuracy.}
\vspace{-4mm}
\label{tab:ood}
\end{table}

\subsection{More Attention Analysis }

\begin{figure}[t!]
  \centering
\includegraphics[width=0.32\textwidth]{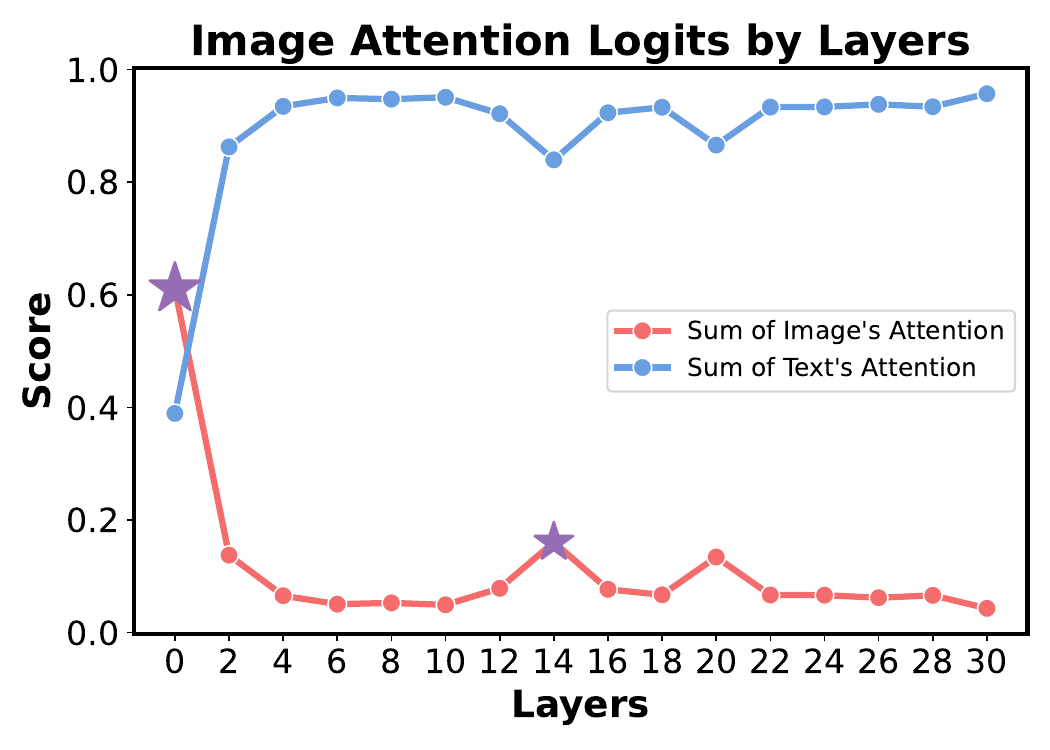}
  \caption{The variance of image token's attention scores through the layers in Cont\_B benchmark.}
  \label{fig:contbattn}
\end{figure}

\begin{figure}[t!]
  \centering
\includegraphics[width=0.32\textwidth]{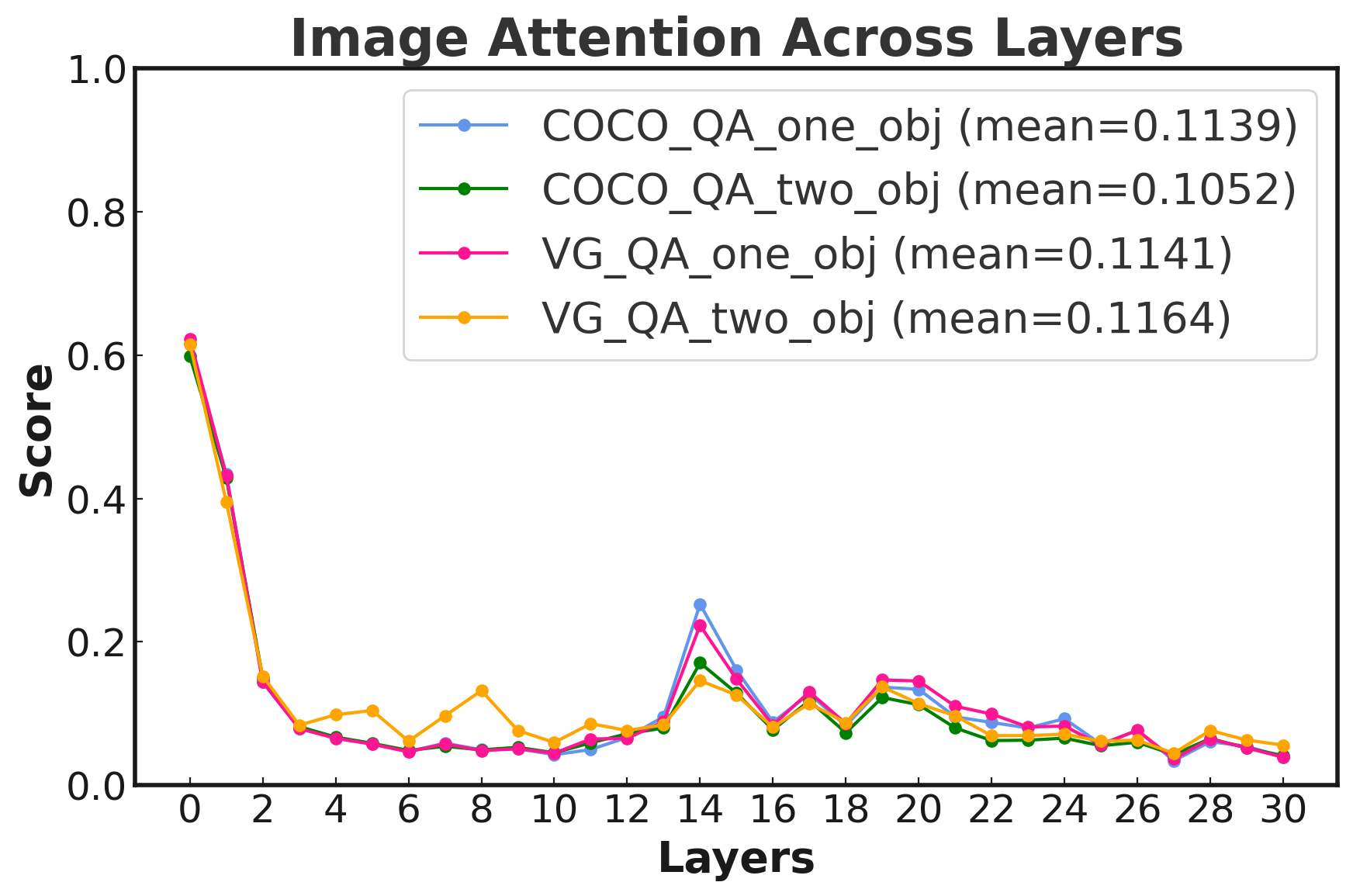}
  \caption{Image attention on COCO and VG dataset.}
  \label{fig:cocovgattn}
\end{figure}

\paragraph{The sparcity}
\label{appdix:analysis}
To determine whether our observation about the sparse attention pattern holds across benchmarks, we examine the attention patterns in other subsets of WhatsUp. Figure~\ref{fig:contbattn} presents the attention pattern for Cont\_B, while Figure~\ref{fig:cocovgattn} illustrates the attention pattern for real images. Consistently, we observe that image attention remains sparse across all subsets.

\paragraph{AUROC Analysis of attention score and confidence}

\begin{figure}[t!]
  \centering
\includegraphics[width=0.4\textwidth]{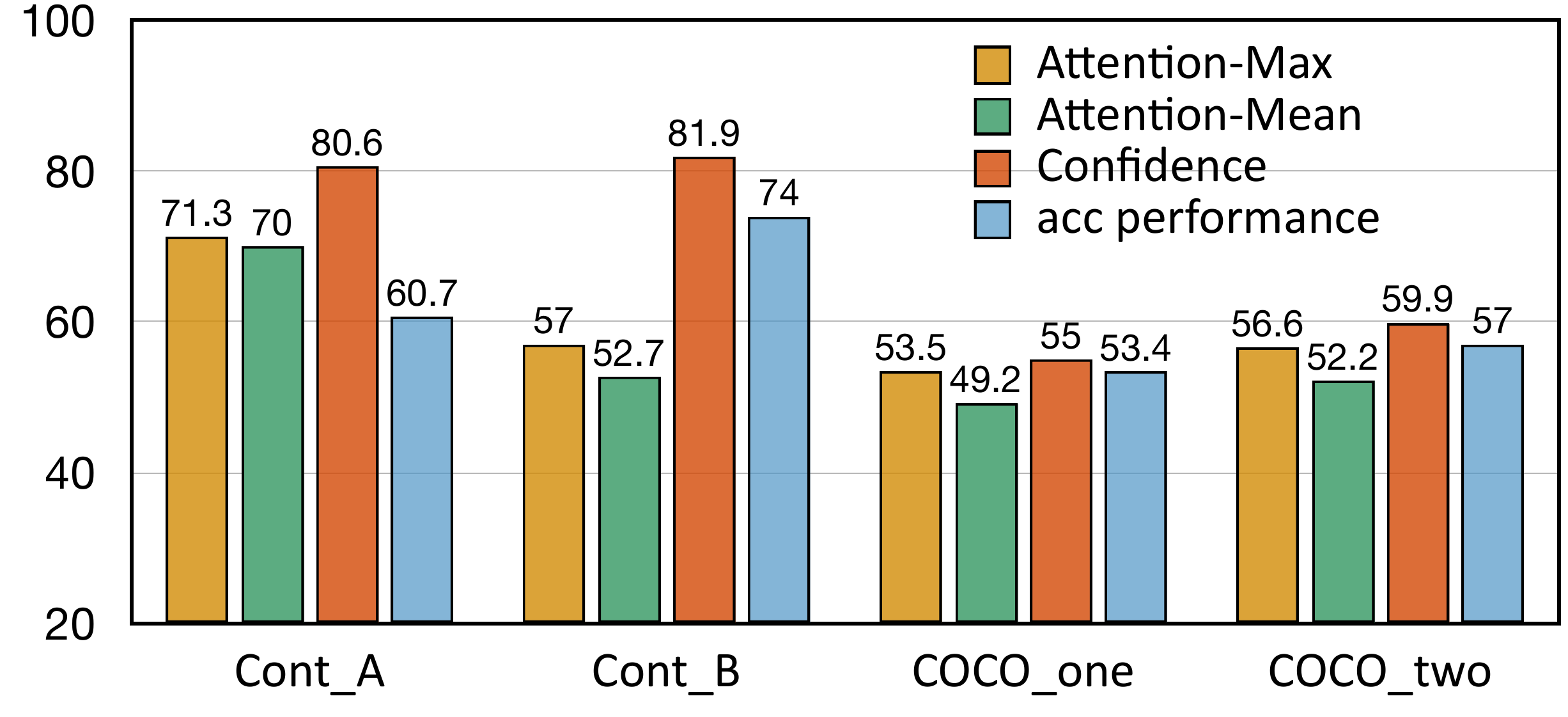}
  \caption{AUROC of attention scores in 17th layer relative to model's confidence. }
  \label{fig:auroc}
\end{figure}
we conduct a calibration experiment using two statistical approaches. The first approach sums the attention scores assigned to the image as a metric, while the second extracts the highest attention score among the image tokens. These metrics are then evaluated to assess their effectiveness in distinguishing between correct and incorrect generations.
However, as shown in Figure~\ref{fig:auroc}, the AUROC score using attention (the yellow and green bar) is consistently lower than the AUROC score of the model's self-confidence (the yellow bar), which we measure by the probability of the output tokens. Additionally, we observe that the maximum attention score provides better calibration than the average attention score, suggesting that key information aligns more closely with maximal attention values. This suggests that the assumption ``\textit{the more attention the model pays to the image, the more accurate the results}" holds only partially true.

\subsection{AUROC anlysis of the overlap between YOLO and the attention scores}
\label{appdix: overlap}
\begin{figure*}[t!]
  \centering
\includegraphics[width=1.0\textwidth]{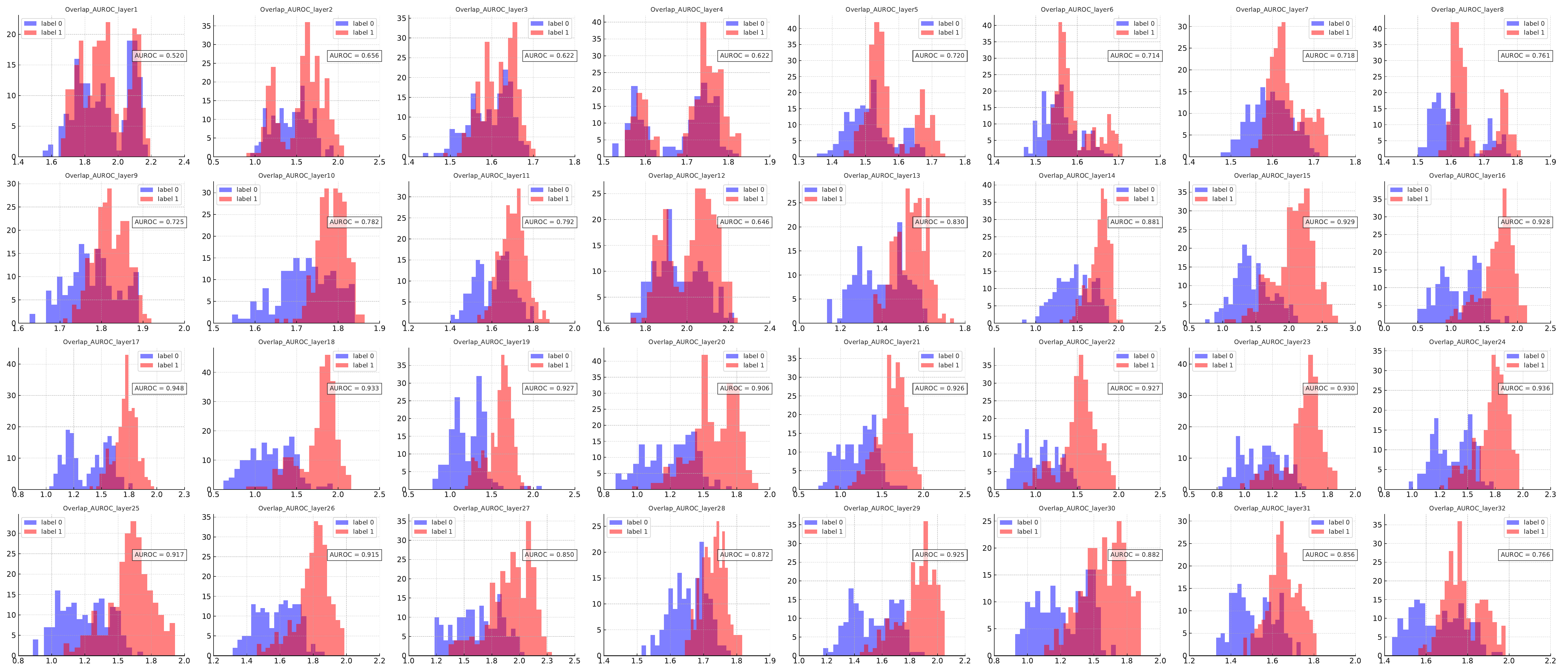}
  \caption{AUROC of the overlap between
YOLO annotations and the attention patterns on all layers in Cont\_A benchmark.}
  \label{fig:overlap_all_layers}
\end{figure*}

\begin{figure*}[t!]
  \centering
\includegraphics[width=1.0\textwidth]{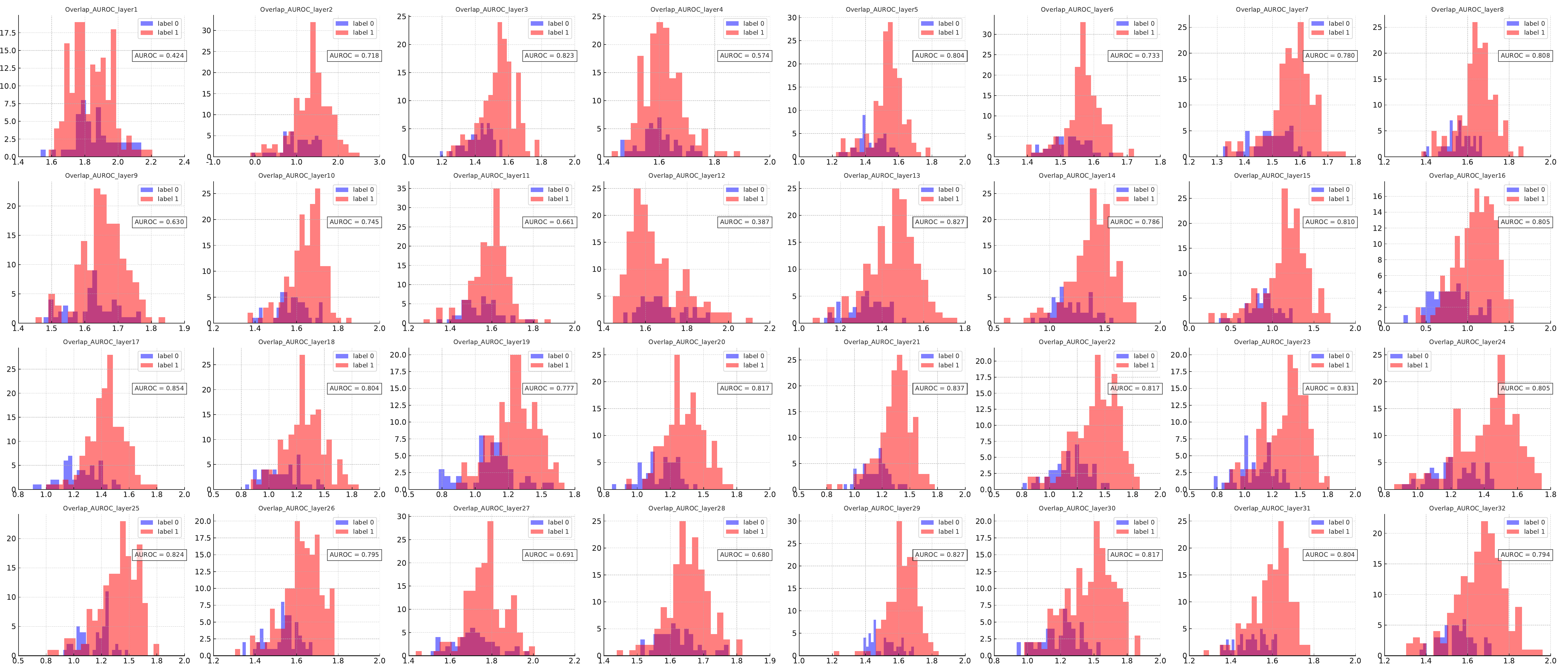}
  \caption{AUROC of the overlap between
YOLO annotations and the attention patterns on all layers in Cont\_B benchmark.}
  \label{fig:overlap_cb}
\end{figure*}

We present the AUROC of the overlap between YOLO annotations and attention scores on the Controlled\_A dataset in Figure~\ref{fig:overlap_all_layers} and Figure~\ref{fig:overlap_cb}. The results demonstrate that the AUROC is remarkably high in the middle-to-high layers, indicating that the attention pattern can serve as a reliable metric for detecting factual errors.

\begin{figure*}[t!]
  \centering
\includegraphics[width=1.0\textwidth]{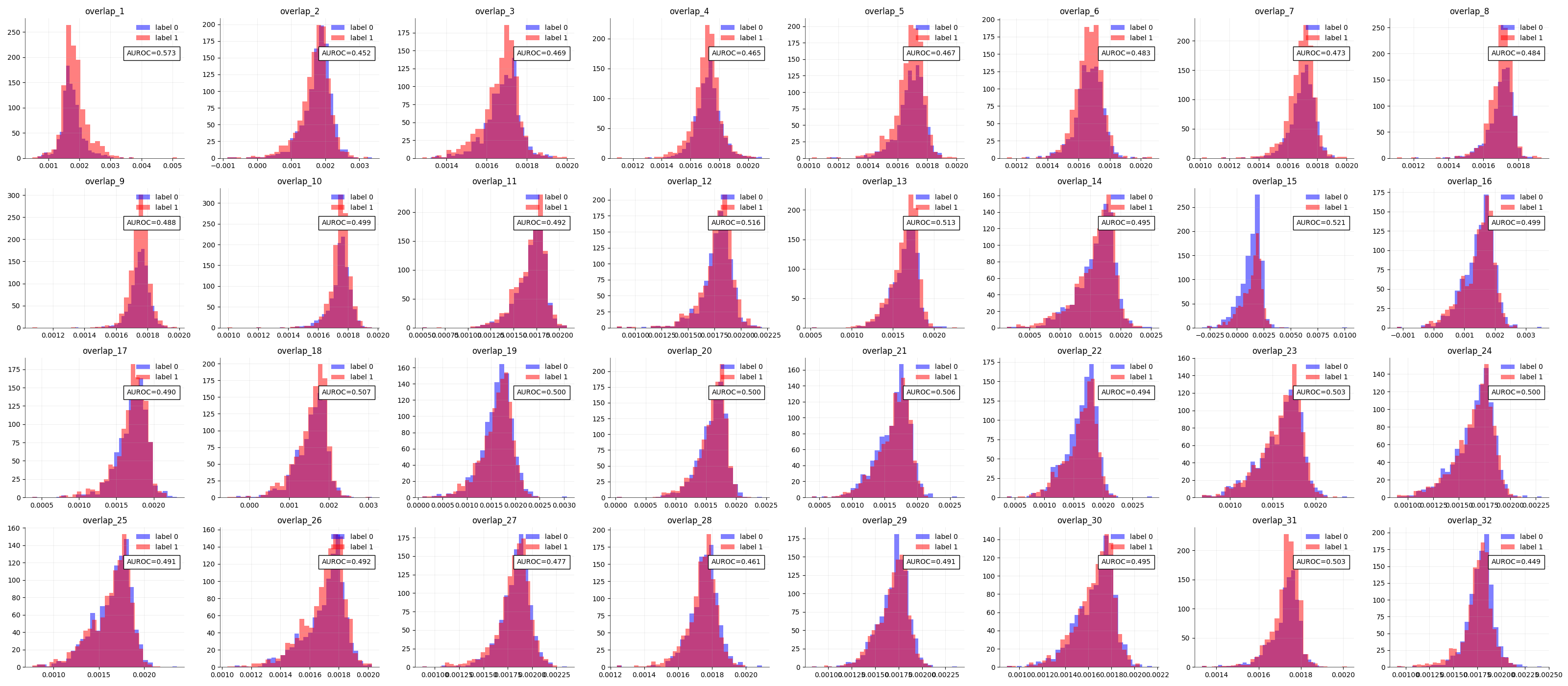}
  \caption{AUROC of the overlap between
YOLO annotations and the attention patterns on all layers in COCO\_one benchmark.}
  \label{fig:coco_one}
\end{figure*}

\begin{figure*}[t!]
  \centering
\includegraphics[width=1.0\textwidth]{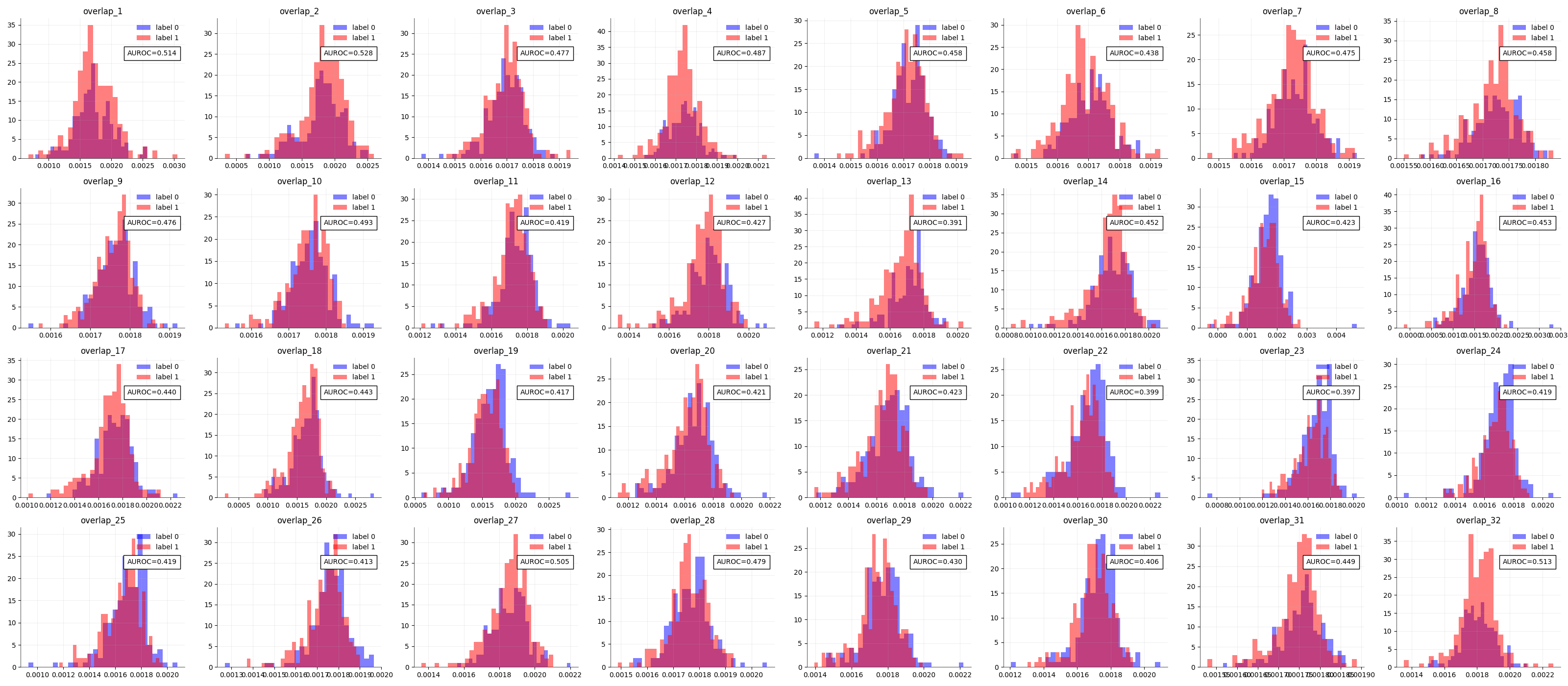}
  \caption{AUROC of the overlap between
YOLO annotations and the attention patterns on all layers in COCO\_two benchmark.}
  \label{fig:coco_two}
\end{figure*}

\begin{figure*}[t!]
  \centering
\includegraphics[width=1.0\textwidth]{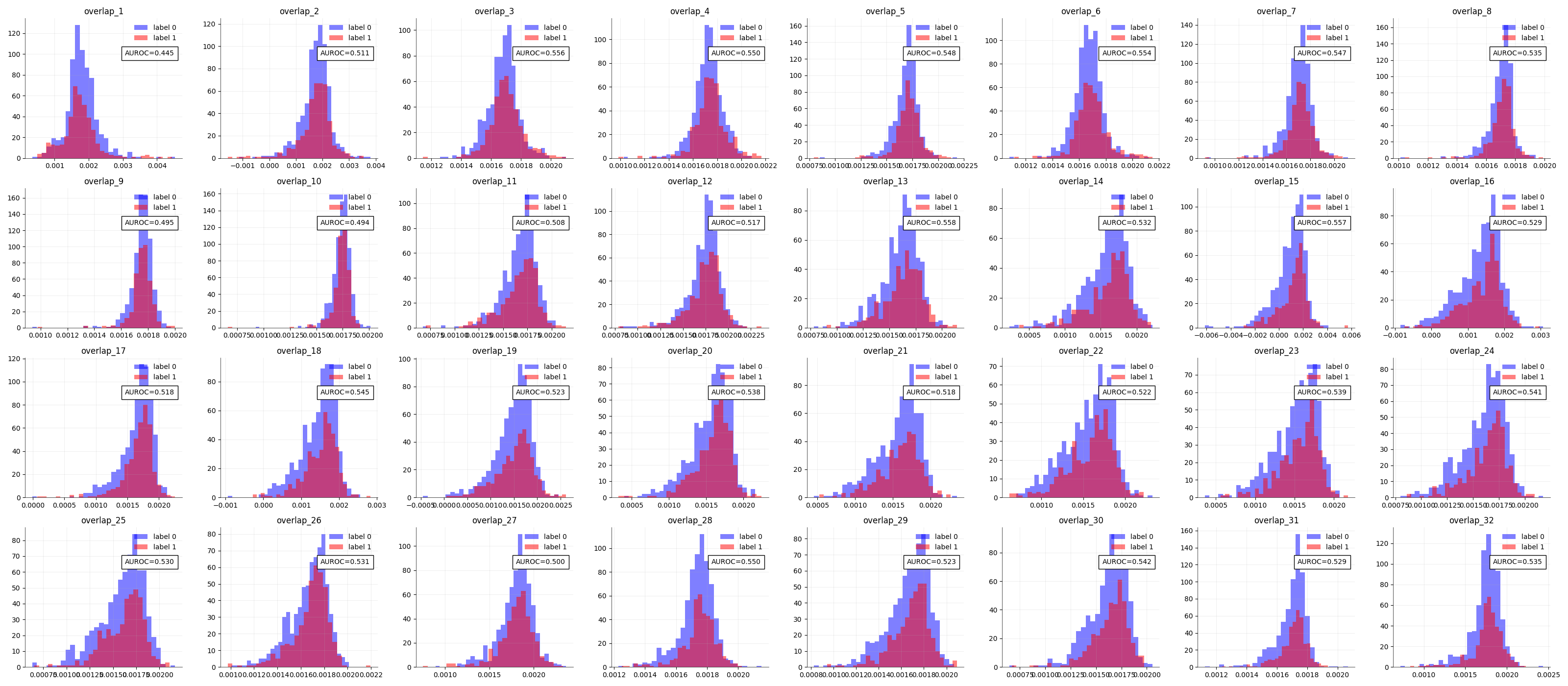}
  \caption{AUROC of the overlap between
YOLO annotations and the attention patterns on all layers in VG\_one benchmark.}
  \label{fig:vg_one}
\end{figure*}

\begin{figure*}[t!]
  \centering
\includegraphics[width=1.0\textwidth]{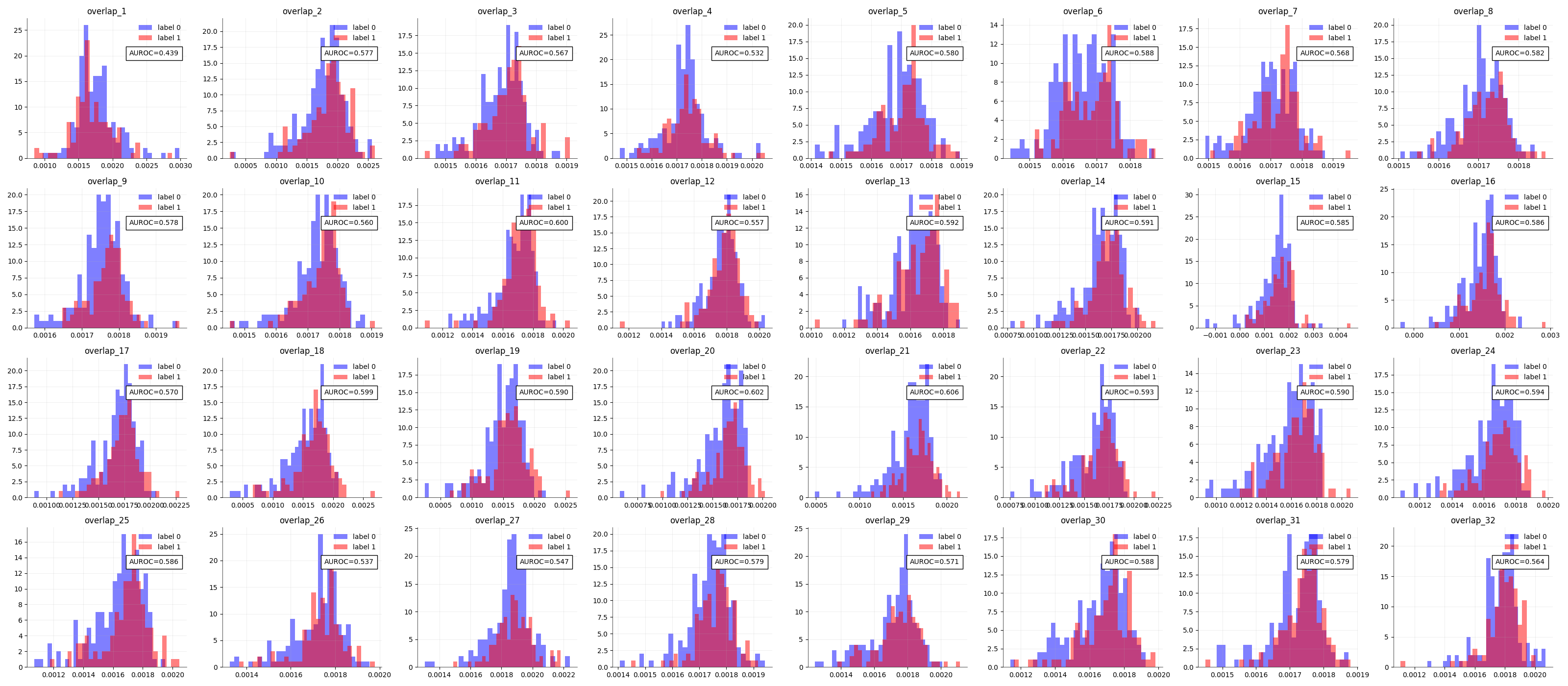}
  \caption{AUROC of the overlap between
YOLO annotations and the attention patterns on all layers in VG\_two benchmark.}
  \label{fig:vg_two}
\end{figure*}

\begin{figure*}[t!]
  \centering
\includegraphics[width=0.9\textwidth]{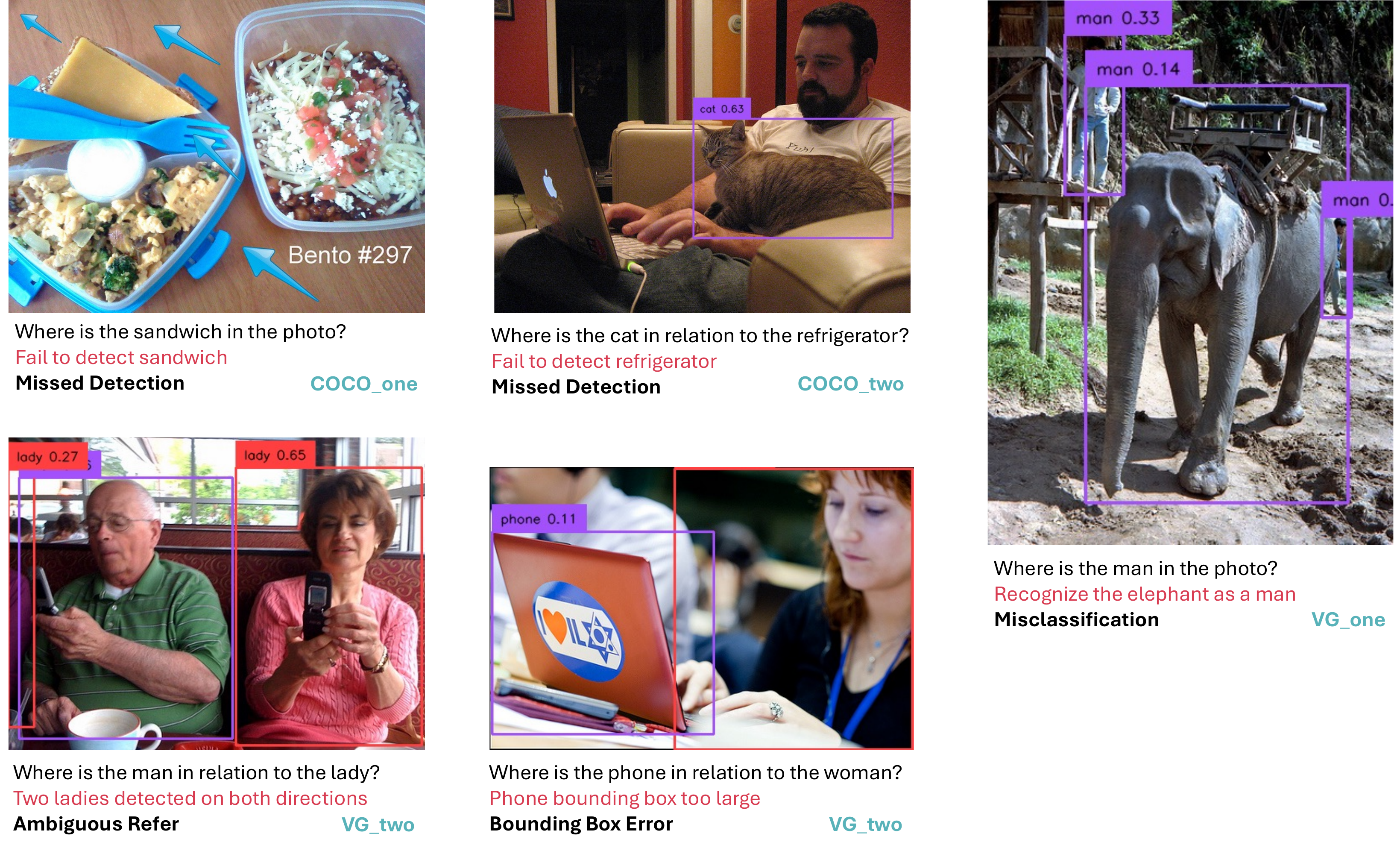}
  \caption{Error cases in YOLO annotation. Under each image we present the related question, the error in YOLO annotation and its type, and the source where the image and question comes from.
  }
  \label{fig:case_study}
\end{figure*}

However, in some cases, this relationship is not as apparent, as shown in Figures~\ref{fig:coco_one}, \ref{fig:coco_two}, \ref{fig:vg_one}, and \ref{fig:vg_two}. In these instances, the AUROC remains relatively modest, which can largely be attributed to errors in YOLO’s annotations. We identified four primary error types leading to mismatches between YOLO annotations and actual object instances: Missed Detection, Misclassification, Bounding Box Error, and Ambiguous Refer. These error patterns are illustrated in Figure~\ref{fig:case_study}.

\clearpage

\clearpage

\subsection{What changes after our intervention?}

\paragraph{How the confidence changes by using different \textbf{$\alpha$}? }To illustrate the impact of coefficients greater or less than 1 on different relationships, Figure~\ref{fig:conf} shows accuracy and confidence variance for various ground-truth relationships. Results indicate that for familiar relationships like ``\textit{left}" (red) and ``\textit{right}" (blue), coefficients greater than 1 boost accuracy and confidence. Conversely, for less familiar relationships like ``\textit{under}" (green) and ``\textit{on}" (yellow), coefficients less than 1 improve accuracy and confidence.

\begin{figure*}[t!]
  \centering
  \hspace*{-0.1\textwidth} 
  \begin{minipage}{0.29\textwidth}
    \centering
    \includegraphics[width=\textwidth]{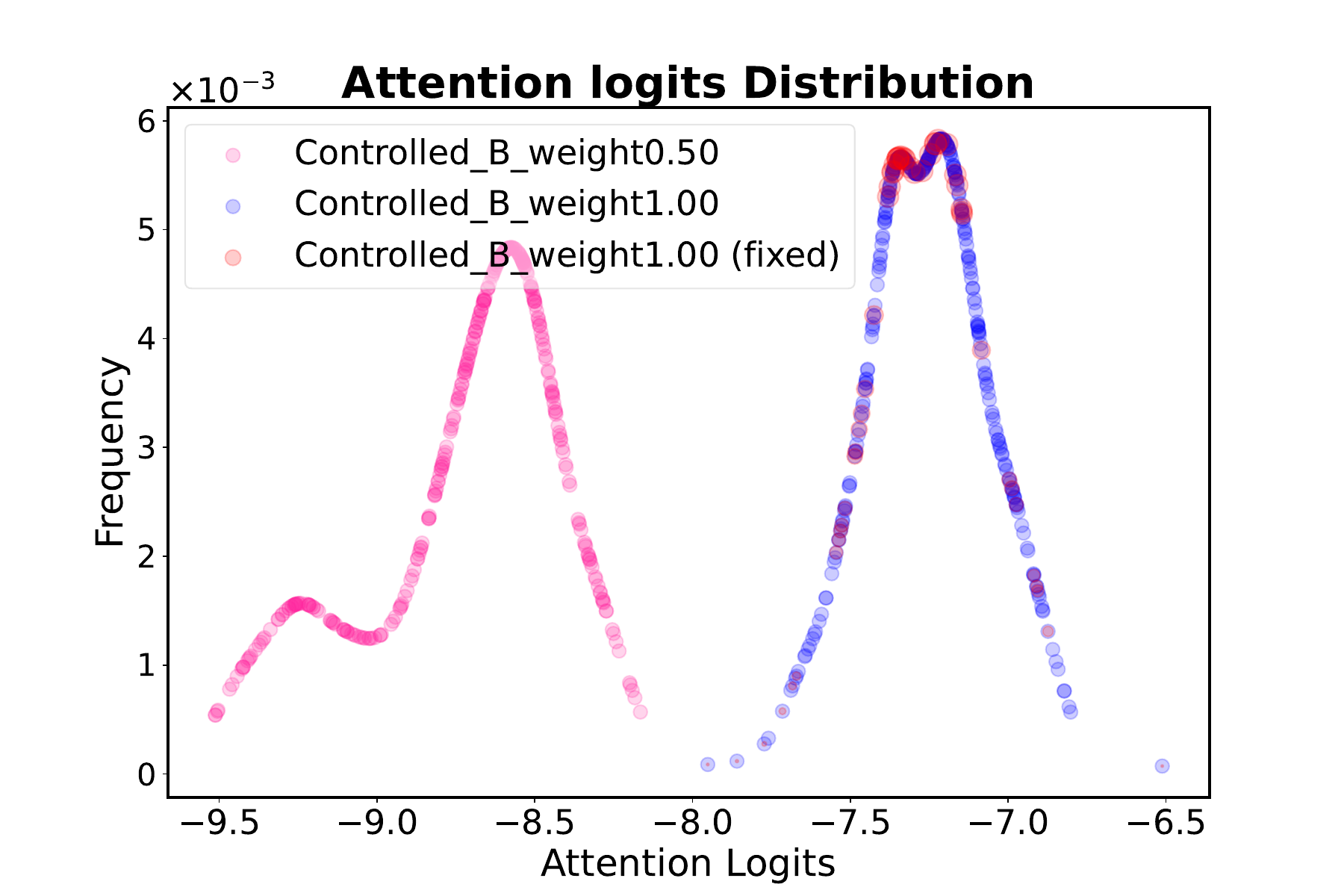}
  \end{minipage}%
  \hspace{-0.035\textwidth} 
  \begin{minipage}{0.29\textwidth}
    \centering
    \includegraphics[width=\textwidth]{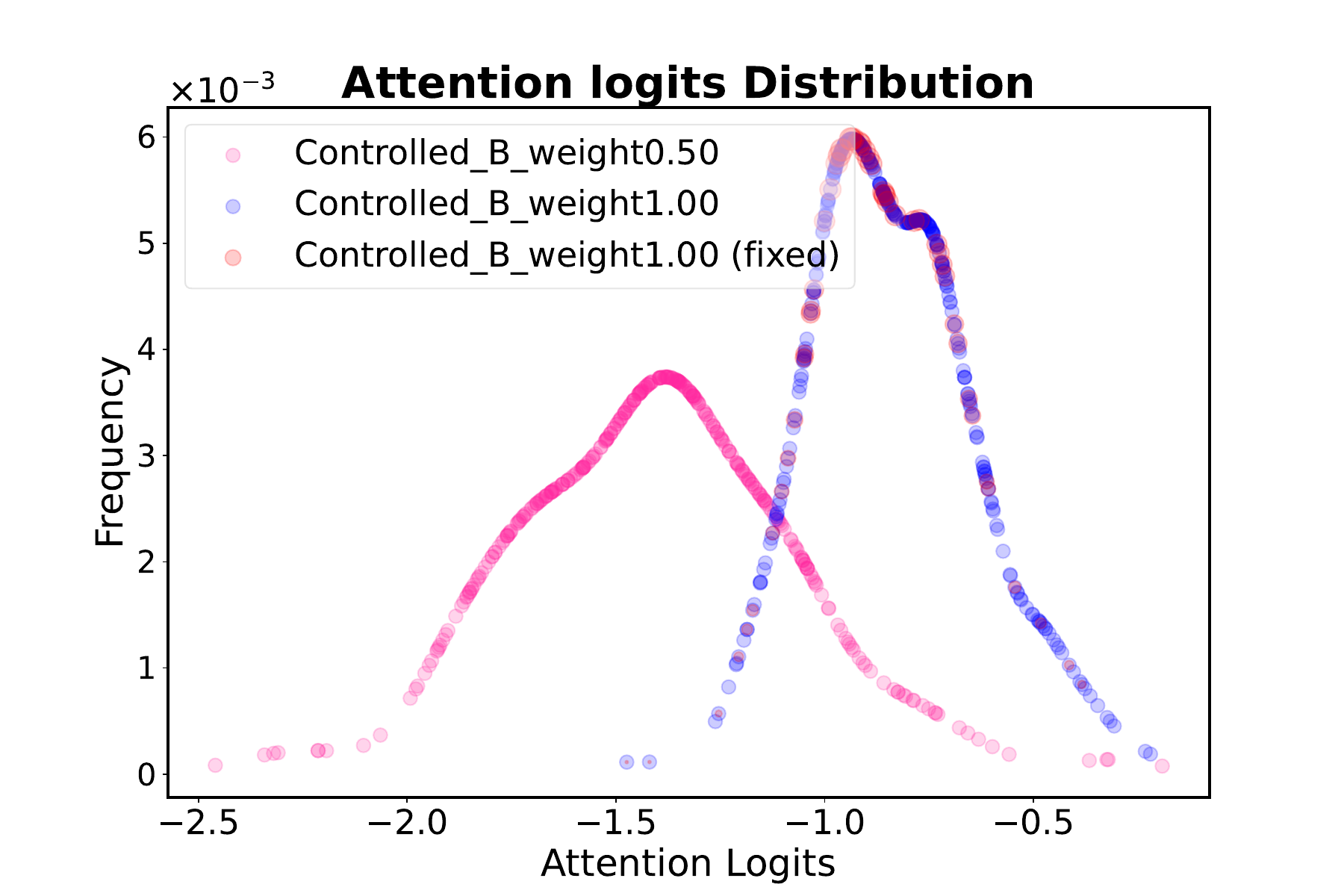}
  \end{minipage}%
  \hspace{-0.035\textwidth} 
  \begin{minipage}{0.29\textwidth}
    \centering
    \includegraphics[width=\textwidth]{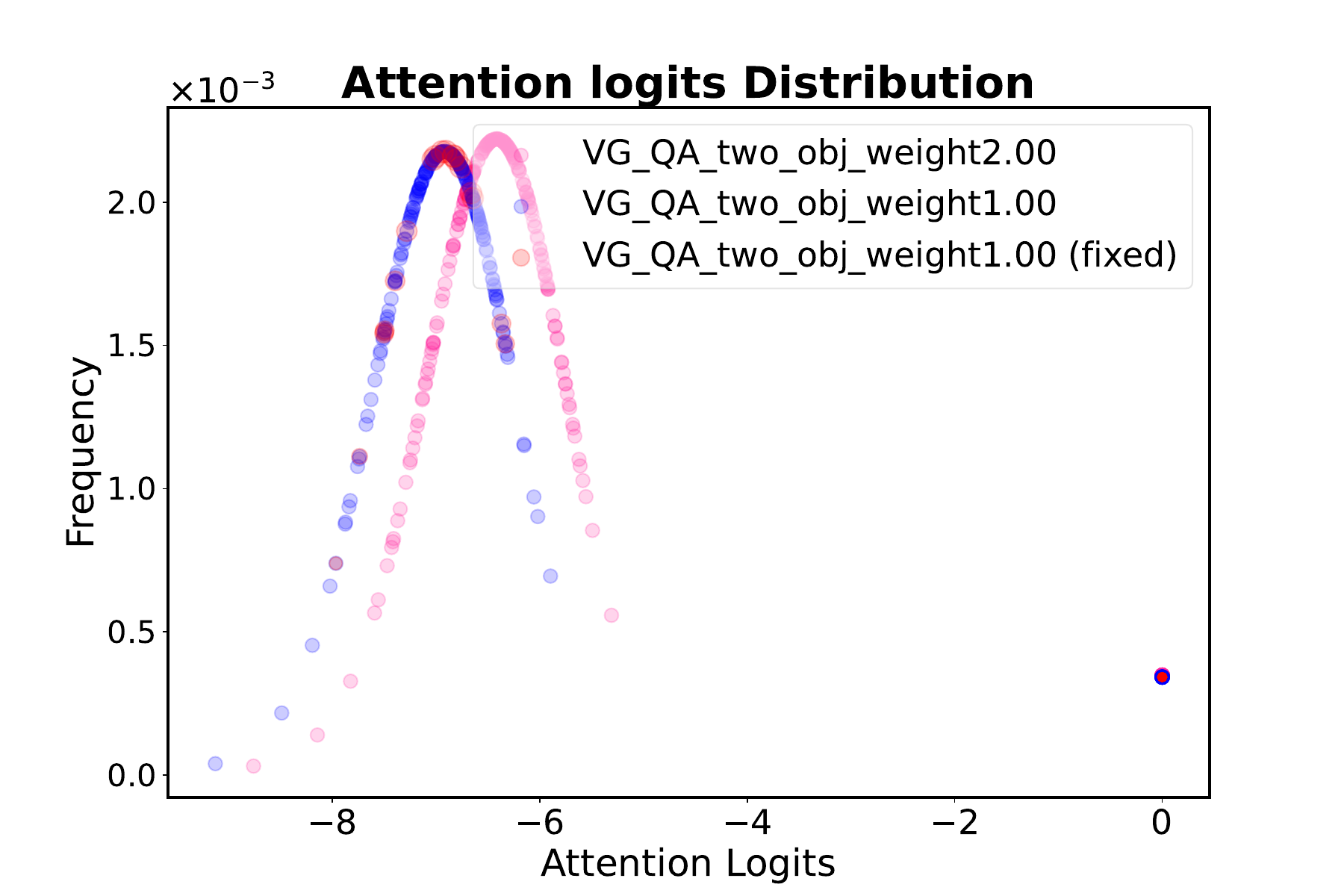}
  \end{minipage}%
  \hspace{-0.035\textwidth} 
  \begin{minipage}{0.29\textwidth}
    \centering
    \includegraphics[width=\textwidth]{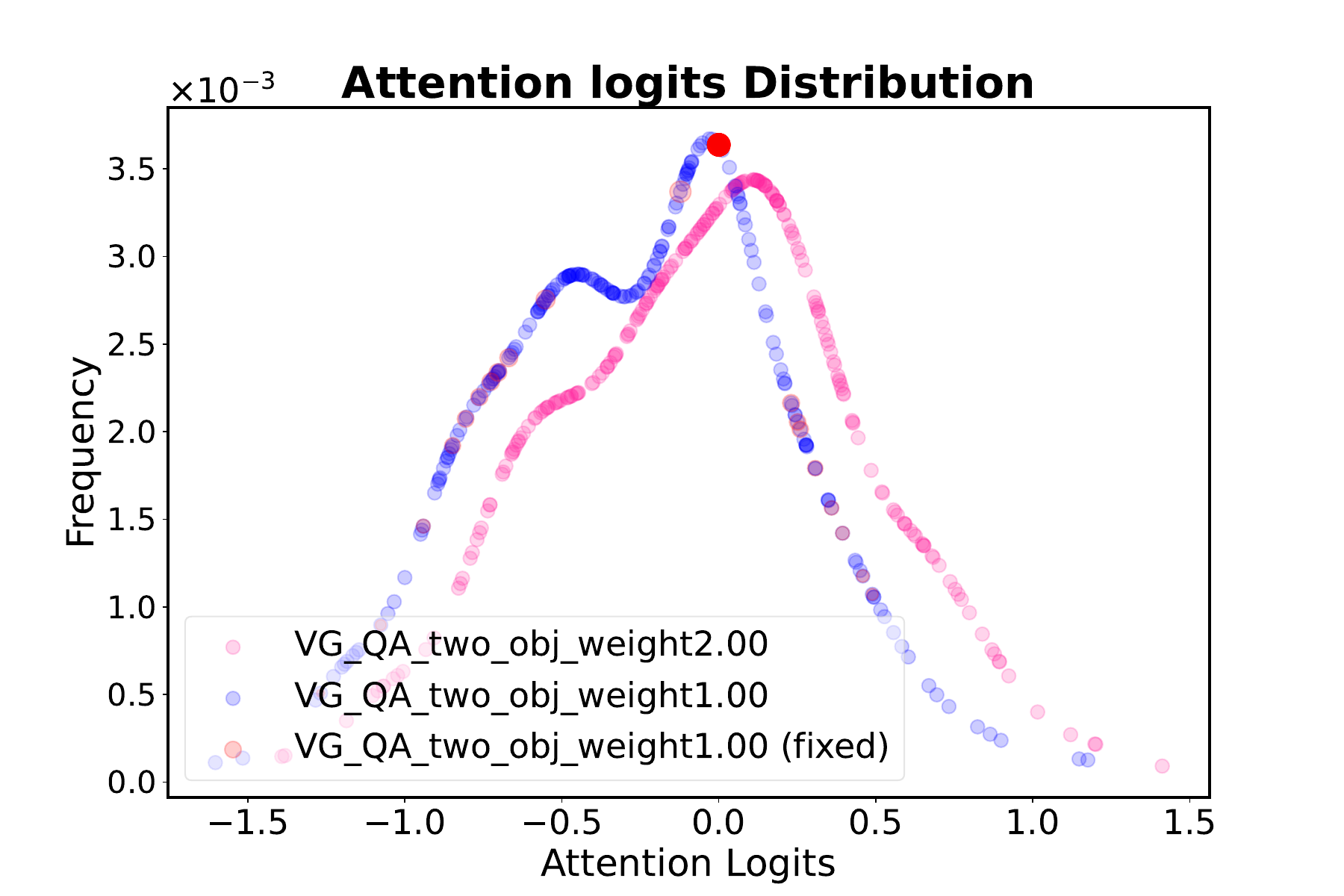}
  \end{minipage}
  \hspace*{-0.1\textwidth} 
  \caption{Attention logit distribution before (blue) and after (pink) intervention in the 14th layer (a randomly chosen middle layer). From left to right, the plots represent: mean and max attention values across heads for Cont\_A, and mean and max attention for Cont\_B. Red dots mark cases corrected by our intervention. We could see that $\alpha$ of 0.5 shifts the line left, while $\alpha$ of 2 shifts it right.} 
  \label{fig:attention-logit-change}
\end{figure*}

\paragraph{How do the absolute values of attention scores vary before and after intervention?} Figure~\ref{fig:attention-logit-change} visualizes the attention logits for two-option Cont\_A ($\alpha$=0.5) and six-option VG ($\alpha$=2). An $\alpha$ of 0.5 decreases the absolute value of logits across layers, while an $\alpha$ of 2 increases them as layers progress. This indicates that an $\alpha$ larger than 1 could strengthen the model's orginal attention pattern.

\subsection{Prompt Sensitivity Analysis}
To assess the robustness of our method, we varied the number of options in prompts for specific WhatsUp subsets. For Cont\_A and Cont\_B, we reduce the number of options to two, simplifying the task. Conversely, for COCO subsets, we increase the options to six, making it more challenging.
Table \ref{tab:change_prompt_res} shows that our \scal method maintains consistent performance across all cases, demonstrating robustness.

\subsection{Reverse curse?}

Additionally, we observe that the model exhibits a ``reverse curse" phenomenon similar to that has been seen in language models~\citep{berglund2023reversal}.

\begin{figure*}[!th]
\centering
\begin{minipage}{0.65\linewidth}
    \centering
    \normalsize
    \renewcommand{\arraystretch}{1.4} 
    \resizebox{1\linewidth}{!}{
    \begin{tabular}{llllllllll}
    \toprule
    \multirow{2}{*}{\textbf{Model}} & \multicolumn{3}{c}{\textbf{Cont\_A}} & \multicolumn{3}{c}{\textbf{Cont\_B}} & \textbf{Coco\_one} & \textbf{Coco\_two} \\
    \cmidrule(lr){2-9}
      & \textbf{Acc} & \textbf{P Acc} & \textbf{S Acc} & \textbf{Acc} & \textbf{P Acc} & \textbf{S Acc} & \textbf{Acc} & \textbf{Acc} \\
    \midrule
    LLaVA-1.5 & 76.4 & 43.0 & 4.8 & 74.6 & 41.0 & 1.2 & 30.8 & 42.6 \\
    +Ours & 86.4\normup{10.0} & 61.2\normup{18.2} & 27.9\normup{23.1} & 87.8\normup{13.2} & 59.3\normup{18.3} & 22.0\normup{20.8} & 36.0\normup{5.2} & 48.6\normup{7.3} \\
    \midrule
    Best $\alpha$ & \multicolumn{3}{c}{\cellcolor{scal}$0.5$} & \multicolumn{3}{c}{\cellcolor{scal}$0.5$} & \cellcolor{adaptive}$2$ & \cellcolor{adaptive}$2$ \\
    \bottomrule
    \end{tabular}}
    \caption{
    Results on WhatsUp (Metrics in $\times 10^{-2}$) when changing prompts for other option numbers. Arrows show improvement over greedy decoding.}
    \label{tab:change_prompt_res}
\end{minipage}%
\hfill
\begin{minipage}{0.34\linewidth}
    \centering
    \includegraphics[width=0.9\linewidth]{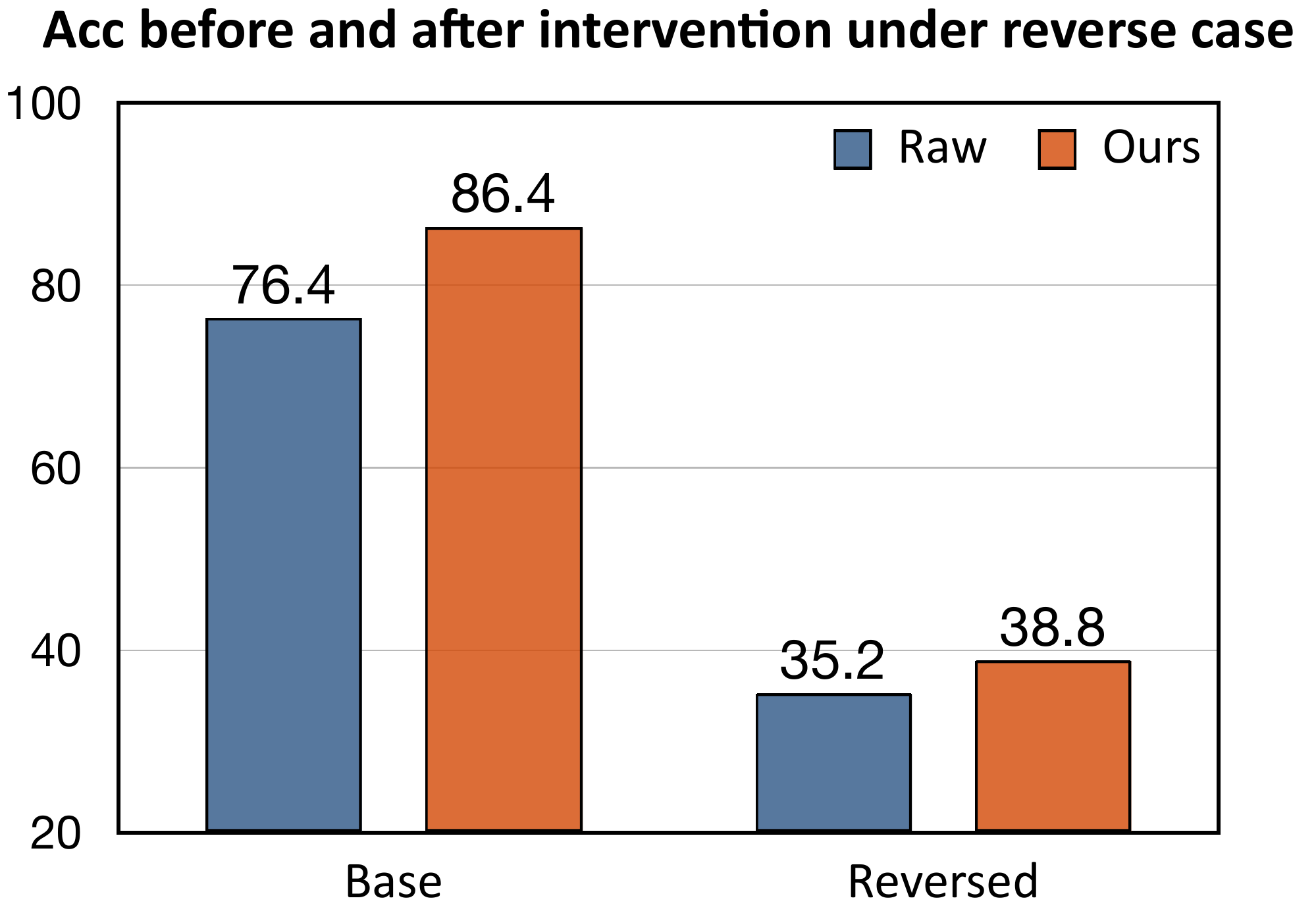}
    \caption{Performance comparison before and after \scal intervention ($\alpha=0.5$). }
    \label{fig:performance-comparison-before-after-scalingvis-reverse-curse}
\end{minipage}
\end{figure*}

When we reverse the order of the entities in Cont\_A (e.g., asking the model,
``\textit{Where is the armchair in relation to the beer bottle?}" instead of ``\textit{Where is the beer bottle in relation to the armchair?}"), there is a significant drop in performance.
As shown in Figure~\ref{fig:performance-comparison-before-after-scalingvis-reverse-curse}, the model's performance declines dramatically from a high score to an exceptionally low one.
This reveals that the existing \vlm's attention pattern and generation results could be significantly impacted by the prompt. Our methods, however, consistently improve the model's performance. It suggests that adaptively intervening the attention score is a generalizable method for different prompts. 

\subsection{Efficiency}
We evaluate inference times for different decoding methods in Table~\ref{tab:efficiency}. ScalingVis introduces only negligible additional computation time compared to the baseline (greedy decoding). However, due to the need for computing the threshold, AdaptVis incurs higher computational overhead.
\begin{table}[t!] 
\centering
\small
\resizebox{\linewidth}{!}{ 
\begin{tabular}{lccc}
\toprule
\textbf{} & \textbf{Baseline} & \textbf{ScalingVis} & \textbf{AdaptVis} \\
\midrule
Time (set baseline to 1) & 1.00 & 1.02 & 1.64 \\
\bottomrule
\end{tabular}
}
\caption{Inference-time statistics. It shows that ScalingVis introduces negligible computation compared to the baseline.}
\label{tab:efficiency}
\end{table}

\subsection{Results on More Benchmarks}
We further evaluate our method on several Question Answering benchmarks, including POPE~\citep{li2023evaluating}, GQA~\citep{hudson2019gqa}, and VQAv2~\citep{antol2015vqa}. As shown in Table~\ref{tab:qares}, our method consistently outperforms the baseline across all benchmarks, demonstrating its strong generalization capability in the general Question Answering setting.
\begin{table}[t!] 
\centering
\small
\resizebox{1.0 \linewidth}{!}{
\begin{tabular}{lcccccc}
\toprule

 \multirow{1}{*}{\textbf{Dataset}} & \textbf{POPE-A} & \textbf{POPE-P} & \textbf{POPE-R} & \textbf{POPE-AVG}& \textbf{GQA} & \textbf{VQAv2}  \\
\midrule
Baseline & 81.0  & 86.4 & 88.4 & 85.3   & 56.1  & 74.0   \\
ScalingVis & 81.8  & 86.6 & 88.6  & 85.6   & 56.3     & 74.2    \\
\bottomrule
\end{tabular}}
\caption{Results on QA benchmarks. POPE-A is POPE-Adversarial. POPE-P is POPE-Popular. POPE-R is POPE-random. POPE-AVG the is average score of the three subsets. }
\label{tab:qares}
\end{table}

\section{URLs of Code and Data}

We provide the code and data at \url{https://anonymous.4open.science/r/AdaptVis-B73C}. 

The running arguments are configured as follows: {\footnotesize\texttt{dataset}} specifies the evaluation dataset (e.g., {\footnotesize\texttt{Controlled\_Images\_A}}); {\footnotesize\texttt{model-name}} selects the model (e.g., {\footnotesize\texttt{llava1.5}}); {\footnotesize\texttt{method}} determines the evaluation approach ({\footnotesize\texttt{scaling\_vis}} or {\footnotesize\texttt{adapt\_vis}}). For {\footnotesize\texttt{scaling\_vis}}, {\footnotesize\texttt{weight1}} can be set to values in $[0.5, 0.8, 1.2, 1.5, 2.0]$. For {\footnotesize\texttt{adapt\_vis}}, \texttt{weight1} ranges in $[0.5, 0.8]$ and {\footnotesize\texttt{weight2}} in $[1.2, 1.5, 2.0]$, with {\footnotesize\texttt{threshold}} controlling the adaptation sensitivity. The {\footnotesize\texttt{--download}} flag enables automatic dataset downloading.

\end{document}